\documentclass[sigconf,nonacm]{acmart}
\usepackage{enumitem}
\usepackage{multirow}
\usepackage{algorithm}
\usepackage{algorithmic}
\usepackage{amsthm}
\newtheoremstyle{flushleftthm}
  {6pt}           
  {6pt}           
  {\itshape}      
  {0pt}           
  {\bfseries}     
  {.}
  {0.5em}         
  {}

\theoremstyle{flushleftthm}
\newtheorem{theorem}{Theorem}
\newtheorem{proposition}{Proposition}

\usepackage{amsmath}
\usepackage{bm}

\AtBeginDocument{%
  }

\setcopyright{acmlicensed}
\copyrightyear{2027}
\acmYear{2027}
\acmDOI{XXXXXXX.XXXXXXX}
\acmConference[WWW '27]{ACM Web Conference 2027}{May 10--14, 2027}{Dublin, Ireland}
\acmISBN{XXX-X-XXXX-XXXX-X/XXXX/XX}

\begin{document}
\newcommand{\CARVE}{\textsc{CARVE}}
\newcommand{\CoStratU}{\textsc{CoStrat-U}}
\newcommand{\Rphi}{\mathcal R_{\varphi}}

\makeatletter
\renewcommand{\proofname}{Proof}
\renewenvironment{proof}[1][\proofname]{%
  \par\pushQED{\qed}%
  \normalfont
  \topsep6pt plus 6pt\relax
  \trivlist
  \item[\hskip\labelsep\itshape #1\@addpunct{.}]%
  \ignorespaces
}{%
  \popQED\endtrivlist\@endpefalse
}
\makeatother
\title{When Cooperation Must Change: Semantic Strategy Unlearning in Multi-Agent Reinforcement Learning}


\author{Wenhan Wu}
\affiliation{%
  \institution{School of Computer Science, Wuhan University}
  \city{Wuhan}
  \state{Hubei}
  \country{China}}
\email{wenhanwu@whu.edu.cn}

\author{Linghan Chen}
\affiliation{%
  \institution{School of Computer Science and Information Technology, University of Adelaide}
  \city{Adelaide}
  \state{South Australia}
  \country{Australia}
}
\email{chenlinghan2004@163.com}

\author{Yudong Gao}
\affiliation{%
 \institution{Department of Computer Science and Engineering, Hong Kong University of Science and Technology}
 \country{Hong Kong SAR}}
\email{ygaodj@connect.ust.hk}

\author{Fan Yang}
\affiliation{%
  \institution{School of Electromechanical and Automation, Huaqiao University}
  \city{Xiameng}
  \state{Fujian}
  \country{China}}
\email{25013088022@stu.hqu.edu.cn}

\author{Honglong Chen}
\affiliation{%
  \institution{College of Control Science and Engineering, China University of Petroleum (East China)}
  \city{Qingdao}
  \state{Shandong}
  \country{China}}
\email{chenhl@upc.edu.cn}

\author{Kai Yin}
\affiliation{%
  \institution{School of Computer Science, Wuhan University}
  \city{Wuhan}
  \state{Hubei}
  \country{China}}
\email{yinkai@whu.edu.cn}

\author{Jiawei Jiang}
\affiliation{%
  \institution{School of Computer Science, Central China Normal University}
  \city{Wuhan}
  \state{Hubei}
  \country{China}}
\email{jiawei.jiang@ccnu.edu.cn}

\author{Chuang Hu}
\correspondingauthor
\affiliation{%
  \institution{the Institute for Math \& AI, Wuhan University}
  \city{Wuhan}
  \state{Hubei}
  \country{China}}
\email{handc@whu.edu.cn}

\renewcommand{\shortauthors}{Wu et al.}

\begin{abstract}
    Cooperative multi-agent reinforcement learning (MARL) supports coordinated decision-making in distributed Web services, multi-robot control, and collaborative games through learned coordination conventions. When deployment requirements change, a previously useful convention may become obsolete or unsafe. For example, revised cache-sharing permissions in Web content delivery may require caches to abandon a learned division of storage and serving roles while maintaining efficient delivery. Teams must therefore remove the specified convention while preserving useful cooperation. This is challenging because the same relational pattern can be reproduced by different agents across different histories. Runtime shielding leaves the underlying coordination preference intact and can impair task progress, while existing reinforcement-unlearning and trajectory-deletion methods do not directly target role-invariant conventions. This paper formalizes this problem as semantic coordination-strategy unlearning and proposes CARVE. CARVE uses causal coordination attribution to identify influential interactions and retain-aware subspace updates to limit interference with unrelated behaviors. Counterfactual re-coordination helps agents learn productive alternatives, while adversarial auditing tests whether the target convention can still be recovered. Accepted updates are merged into decentralized actors for direct deployment. Evaluation across five cooperative MARL benchmarks shows that CARVE reduces mean nominal convention-completion and worst-case elicitation rates by 91.4\% and 83.3\%, respectively. The median utility loss is 5.9\%, and measured inference latency matches that of the original policies. 
\end{abstract}



\maketitle

\section{Introduction}
Cooperative multi-agent reinforcement learning (MARL) enables agents to pursue shared objectives in distributed Web services, multi-robot systems, and collaborative games. Its effectiveness often depends on learned coordination conventions: distributed edge caches coordinate content placement and retrieve objects from neighboring caches to reduce content-delivery latency~\citep{chang2022cooperative}, allied units focus fire in StarCraft micromanagement~\citep{samvelyan2019starcraft,ellis2023smacv2}, partners divide cooking and delivery roles in Overcooked~\citep{carroll2019utility,ruhdorfer2024overcooked}, and robots negotiate passage through narrow corridors~\citep{bettini2022vmas,papoudakis2020benchmarking}. These conventions simplify joint decision-making and can contribute substantially to policy return. Once formed, however, they also become behavioral patterns that deployed systems may repeatedly reproduce. Removing an obsolete or unsafe coordination pattern while preserving service efficiency and effectiveness is therefore an important problem for reliable multi-agent deployment.

In real-world settings, changing deployment requirements may render some coordination conventions no longer permissible. For example, in a cooperative Web caching service, controllers may learn a division of roles in which one cache leaves an object uncached locally while another retains it and serves subsequent peer fetches. This convention can improve storage utilization and reduce origin requests. If updated service rules prohibit sharing a designated content class across an administrative boundary, the team must abandon conventions that depend on the revoked sharing relation while continuing to serve requests through local caches, permitted peers, or the origin. Such a request targets a specific mode of cooperation, not the entire task and not one particular agent. Removing the target convention therefore requires targeted intervention: the system should abandon the specified convention while retaining the coordination skills needed to make progress in other ways. We call this problem \emph{semantic coordination-strategy unlearning}.

A natural response is to prohibit the target convention at execution time to satisfy the semantic unlearning objective. In the caching example, a runtime rule can reject disallowed peer fetches, but learned controllers may continue to omit local copies, increasing origin traffic or delivery latency when those fetches are rejected. Our passage experiment in the Vectorized Multi-Agent Simulator (VMAS), a two-dimensional physics simulator for multi-agent robotic learning~\citep{bettini2022vmas}, illustrates this tradeoff between convention suppression and task performance in a controlled coordination setting. The policy is trained with Multi-Agent Proximal Policy Optimization (MAPPO), which uses centralized value estimation during training while allowing agents to choose actions in a decentralized manner at execution time~\citep{yu2022surprising}. In this task, the policy learns a bottleneck convention: one agent yields or blocks while another agent passes. Starting from a restored checkpoint, we apply an automaton-based runtime shield that intercepts actions completing the forbidden convention~\citep{hammond2021multi,icarte2018using}. Although the shield changes the executed action, the original actor network continues to propose the same action pattern. The agents are repeatedly intervened on near the passage and fail to make efficient progress. As shown in Table~\ref{tab:disable-motivation}, the shield reduces WER from \(0.755\) to \(0.341\), where $\mathrm{WER}=\max_{\nu\in\mathcal V}\Pr_{\pi,\nu}(\tau\models\varphi_f)$ is the worst-case elicitation, computed over held-out or adversarial contexts \(\nu\). Here, \(\tau\) denotes a trajectory, \(\varphi_f\) is a relational temporal formula specifying the forbidden convention, and \(\tau\models\varphi_f\) means that the trajectory realizes an accepting instance of that convention, such as one agent blocking or yielding followed by another agent passing within a bounded time window. Lower WER means the forbidden convention is harder to recover under role permutations, starts, or context changes. However, the task return drops from \(47.5\) to \(29.6\), and normalized task utility falls to \(0.623\). Return is the episodic task reward measuring task performance, while utility normalizes how much of the original performance is retained. Thus, an apparent improvement in a convention-suppression metric can hide a severe loss of performance. The real challenge is not only to block the old convention, but to change the learned coordination behavior while preserving a viable way to complete the task.

This difficulty is not limited to runtime constraints, because a coordination convention is not owned by one agent or one trajectory. Existing reinforcement-unlearning and trajectory-deletion methods do not directly address semantic strategy removal~\citep{ye2023reinforcement,gong2024trajdeleter}. For example, in the caching scenario, the revoked convention concerns the relation between a cache that omits a local copy and a peer that subsequently supplies the object across the restricted boundary. After deleting trajectories involving one cache pair, other nodes may still reproduce the same placement and fetching pattern. Similarly, forgetting focus fire must cover the same tactic across different allied units and enemy targets. We call this property \emph{role invariance}: the target convention is defined by roles and interactions rather than fixed agent identities. Moreover, the influence of the convention may be distributed across shared policies, value networks, communication modules, and agents' responses to teammates. These properties show that fixed agents, trajectories, or state-action pairs are too narrow to serve as the basic unit of such unlearning requests.

These observations motivate a behavioral notion of semantic coordination-strategy unlearning: an edited policy should suppress the specified convention across nominal and eliciting contexts while preserving useful cooperation. We think target conventions should be specified after deployment, because shared team rewards entangle their contributions with useful cooperation, making it difficult to withhold positive credit selectively during training. We therefore study request-driven post-training policy editing, with the accepted update merged into decentralized actors for direct deployment. It provides a concrete comparison target beyond the frequency of forbidden events and includes the preservation of original coordination competence as part of the unlearning requirement. The final deployed policy should satisfy these behavioral requirements without relying on a detachable inference-time prohibition layer.

\begin{table}[t!]
\centering
\begin{tabular}{@{}lrrr@{}}
\toprule
Policy & WER$\downarrow$ & Return$\uparrow$ & Utility$\uparrow$ \\
\midrule
Original checkpoint & 0.755 & 47.5 & 1.000 \\
Runtime shield & 0.341 & 29.6 & 0.623 \\
\CARVE{} (Ours) & 0.340 & 158.2 & 1.000 \\
\bottomrule
\end{tabular}
\caption{Comparison on VMAS Passage from the same restored MAPPO checkpoint. Similar WER values can correspond to different task performance.}
\label{tab:disable-motivation}
\end{table}

To address this MARL problem, we propose \CARVE{}. 
The framework first uses counterfactual action interventions to attribute target-convention risk to agent-time interactions. It then selects parameter subspaces according to the behavior that should be retained and restricts updates to these subspaces, reducing interference with other coordination skills. Counterfactual re-coordination helps agents learn alternative ways to complete the task, avoiding the progress failures caused by merely blocking the original action pattern. Finally, adversarial auditing tests whether the target convention can still be recovered. \CARVE{} connects the semantic scope of an unlearning request with the cooperative behaviors that must be preserved, enabling targeted model updates for semantic coordination-strategy unlearning. Our evaluation spans five representative MARL environments and a set of representative baselines to assess the effectiveness and efficiency of policy unlearning. Compared with runtime shielding, \CARVE{} achieves relative improvements of \(167.8\%\) in average task utility. Unlearning is \(101.92\)--\(584.83\) times faster than original backbone training. Meanwhile, the two methods achieve similar average reductions in WER, indicating that reporting target-convention recoverability alone is insufficient; evaluation must also consider task utility, return retention. Our main contributions are summarized as follows:
\begin{itemize}[leftmargin=*]
    \item We first identify limitations of MARL runtime shield and provide the formalization of role-invariant semantic coordination-strategy unlearning in finite-horizon cooperative Markov games.
    \item We propose \CARVE{}, combining causal interaction attribution, counterfactual re-coordination, and adversarial auditing for efficient subspace updates that remove target strategies while preserving performance.
    \item We evaluate \CARVE{} across five MARL benchmarks through basic and stronger elicitation tests, efficiency measurements, and ablations. Compared with the strongest parameter-editing baseline, \CARVE{} improves unlearning effectiveness by \(16.9\%\).
\end{itemize}

\section{Related Work}

\paragraph{MARL benchmarks and algorithms.}
Reinforcement learning~\citep{arulkumaran2017deep, dai2025reinforcement} is gradually evolving towards multi-agent systems. A large body of cooperative MARL research has developed benchmarks and algorithms for learning coordination. SMAC established StarCraft micromanagement as a cooperative MARL benchmark \citep{samvelyan2019starcraft}; SMACv2 introduced procedural generation and stronger closed-loop evaluation pressure \citep{ellis2023smacv2}. Overcooked-AI studies human-AI coordination \citep{carroll2019utility}, and the Overcooked Generalisation Challenge (OGC) expands evaluation to novel partners and layouts \citep{ruhdorfer2024overcooked}. VMAS provides fast vectorized multi-robot simulation \citep{bettini2022vmas}. RWARE and LBF are sparse-reward coordination benchmarks popularized through systematic cooperative MARL evaluation \citep{papoudakis2020benchmarking}. Melting Pot complements these with test-time social generalization \citep{leibo2021scalable}. Standard backbones include MAPPO \citep{yu2022surprising}, QMIX \citep{rashid2020monotonic}, VDN \citep{sunehag2017value}, MADDPG \citep{lowe2017multi}, and COMA \citep{foerster2018counterfactual}. Our work builds on these established environments and backbones, but changes the evaluation question from whether agents can acquire coordination to whether a trained system can selectively forget it while retaining task competence.

\paragraph{Reinforcement and machine unlearning.}
Machine unlearning~\citep{oesterling2024fair,bourtoule2021machine} is commonly motivated by retraining equivalence: the post-unlearning model should resemble a model trained without the removed data or influence \citep{liang2026federated,wu2025mimir,wu2025zero,wu2026remisvfu,liang2026decoupled} due to the requirements of GDPR~\citep{zaeem2020effect}, CCPA~\citep{bonta2022california} or users' privacy needs~\citep{wu2025aegis,wu2025defending}. Reinforcement unlearning extends this idea to RL environments \citep{ye2023reinforcement}, while TrajDeleter targets offline RL trajectory removal \citep{gong2024trajdeleter}. These settings do not cover a distributed, role-invariant coordination convention that can be re-enacted by different agents.

\paragraph{Policy repair, safe MARL, and temporal logic.}
Another line of work studies how to constrain or repair policies after undesirable behavior has been identified. Temporal logic and reward machines provide automata-based structure for non-Markovian tasks \citep{hammond2021multi,icarte2018using,chen2026system}. Safe MARL methods such as MACPO and MAPPO-Lagrangian optimize constraints during policy improvement \citep{gu2021multi}. PolicyCleanse mitigates RL backdoors \citep{guo2023policycleanse}, and MUTE removes low-value communication \citep{zuo2026mute}; both are relevant conditional baselines for triggered or communication-mediated target strategies. These approaches provide closely related risk–return objectives and are essential baselines for behavioral strategy unlearning. CARVE focuses on request-driven editing of a trained team, using interaction attribution, retain-aware sparse updates, and teacher-guided replacement coordination to suppress role-invariant conventions while preserving cooperative task performance.
\section{Preliminaries}
We study a finite-horizon cooperative Markov game as follows:
\begin{equation}
\mathcal G=(\mathcal N,\mathcal S,\{\mathcal O_i\},\{\mathcal A_i\},P,r,\gamma,H),
\end{equation}
where \(\mathcal N=\{1,\ldots,N\}\) is the set of agents, \(\mathcal S\) is the global state space, \(\mathcal O_i\) and \(\mathcal A_i\) are the observation and action spaces of agent \(i\), \(P\) is the transition kernel, \(r\) is the shared team reward, \(\gamma\) is the discount factor, and \(H\) is the horizon. At time \(t\), agent \(i\) observes \(o_t^i\), acts with \(a_t^i\), and conditions on its local history \(h_t^i\). We use \(\mathbf h_t=(h_t^1,\ldots,h_t^N)\) and \(\mathbf a_t=(a_t^1,\ldots,a_t^N)\) for the joint history and joint action. A decentralized policy factorizes as
\begin{equation}
\boldsymbol\pi_{\theta}(\mathbf a_t\mid\mathbf h_t)
=
\prod_{i=1}^{N}\pi_{\theta_i}(a_t^i\mid h_t^i),
\end{equation}
where \(\theta\) denotes the actor parameters. Parameter sharing is allowed: multiple agents may use the same parameter block while receiving different histories. The return of a policy is
\begin{equation}
J(\pi)=
\mathbb E_{\tau\sim\pi}\left[
\sum_{t=0}^{H-1}\gamma^t r(s_t,\mathbf a_t)
\right],
\end{equation}
where \(\tau\) denotes a complete trajectory.

\subsection{Semantic Requests}

The object to be forgotten is not a data point, one fixed agent, or one exact trajectory. It is a \emph{role-invariant coordination convention}, which can be specified as a relational temporal formula \(\varphi_f\), where \(f\) denotes ``forget.'' Atomic predicates are event labels computed from the state and joint action, such as \(\operatorname{Block}(i,\mathrm{door})\), \(\operatorname{Pass}(j,\mathrm{door})\), or \(\operatorname{Attack}(i,e)\). The variables \(i,j\) range over agents, and \(e\) ranges over task entities, such as a specific enemy in a game. For example, the semantic request ``forget the convention where one agent blocks a door so another passes shortly after'' can be written as
\begin{equation}
\varphi_f=
\Diamond\!\left[
\exists i\neq j:
\operatorname{Block}(i,\mathrm{door})
\land
\Diamond_{\le k}\operatorname{Pass}(j,\mathrm{door})
\right].
\end{equation}
Here \(\Diamond\) means ``eventually'' and \(\Diamond_{\le k}\) means ``within \(k\) steps.'' \(\exists i\neq j\) means that the convention should be removed no matter which concrete agents instantiate the blocker and passer roles. The same notation also covers combat conventions such as focus-fire:
\begin{equation}
\varphi_f=
\Diamond\!\left[
\exists i\neq j,e:
\operatorname{Attack}(i,e)
\land
\Diamond_{\le k}\operatorname{Attack}(j,e)
\right],
\end{equation}
which asks the policy to forget the role-invariant tactic where two allies successively attack the same enemy \(e\).
For monitoring, we compile \(\varphi_f\) into a finite automaton
\begin{equation}
\mathcal A_\varphi=(Q,q_0,\delta,F).
\end{equation}
\(Q\) is the set of monitor states, \(q_0\) is the initial state, \(\delta\) updates the monitor after seeing the current event labels, and \(F\) is the set of accepting states indicating that the forbidden convention has occurred. The product state
\begin{equation}
\bar s_t=(s_t,q_t)
\end{equation}
combines the environment state \(s_t\) with the current monitor state \(q_t\). Thus the monitor turns a history-dependent semantic request into a Markovian signal that can be used by critics, attribution, and evaluation.

\subsection{Unlearning Goal}
Let \(c\in\mathcal C\) denote an evaluation context, such as a layout, initial-state distribution, teammate population, or adversarial partner. The semantic risk of a policy is
\begin{equation}
\Rphi(\pi;c)=
\Pr_{\tau\sim(\mathcal G_c,\pi)}[\tau\models\varphi_f],
\qquad
\Rphi^{\mathrm{rob}}(\pi)=
\sup_{c\in\mathcal C}\Rphi(\pi;c).
\end{equation}
The first term is the probability that the policy executes the forbidden convention in context \(c\). The second is the worst-case risk over the evaluation contexts. Given an original policy \(\pi_0\), an unlearning algorithm returns a new policy \(\pi^-=U(\pi_0,\varphi_f)\). We want \(\pi^-\) to satisfy two behavioral requirements:
\begin{equation}
\Rphi^{\mathrm{rob}}(\pi^-)\le\epsilon_f,
\quad
J_r(\pi^-)\ge J_r(\pi_0)-\delta_J.
\end{equation}
The first condition asks for low residual semantic risk under nominal and adversarial evaluation contexts. The second preserves retained task performance; \(J_r\) is return measured on target-unrelated or retained contexts, and \(\delta_J\) is the allowed loss. We do not require a separately trained reference policy. Instead, the evaluation directly measures whether the forbidden convention occurs or can be elicited again, using semantic monitors under nominal rollouts, role permutations, held-out contexts, and adversarial partners. This distinction is important because a runtime shield may block the final forbidden action while leaving the underlying coordination preference unchanged. The request is intentionally behavioral in approximate unlearning~\citep{xu2024machine,wang2024machine,li2025machine}, which does not claim that every parameter-level trace of the original strategy disappears. Instead, they ask whether the delivered policy acts like one that no longer relies on the forbidden convention, while keeping the rest of its cooperative competence.

\section{System Design: \CARVE{}}

\begin{figure*}
    \centering
    \includegraphics[width=0.95\linewidth]{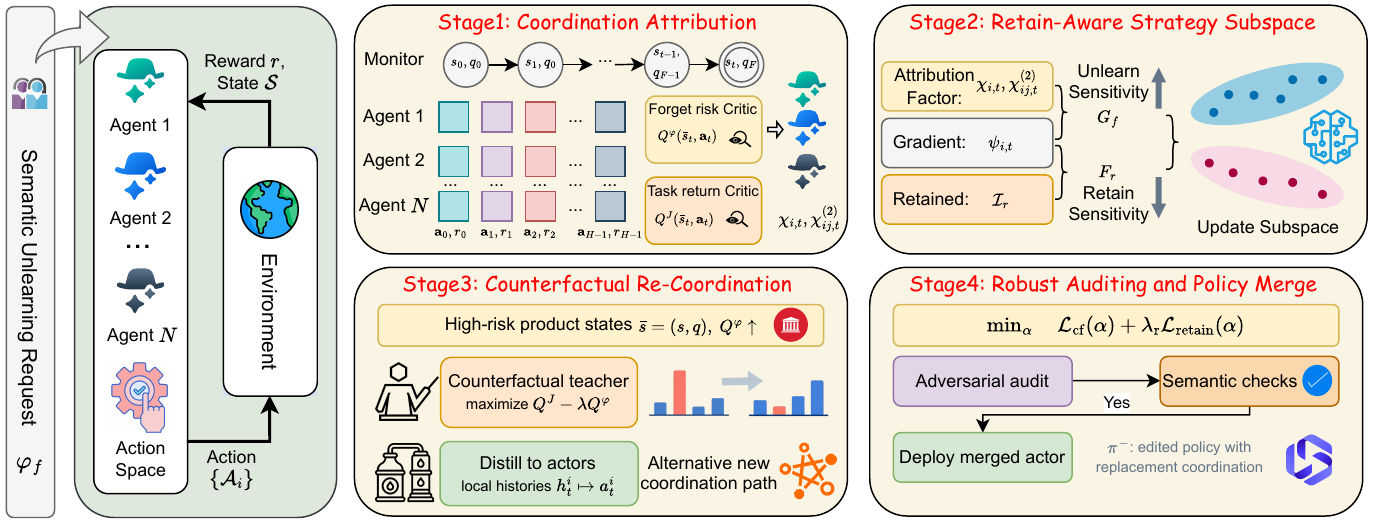}
    \caption{Overview of \CARVE{}. The framework takes an original policy, rollout data or simulator access, and a semantic request \(\varphi_f\), and returns an edited policy \(\pi^-\). \CARVE{} first attributes forbidden-convention risk to agent-time interactions, then selects retain-aware sparse edit directions, constructs replacement coordination through a counterfactual teacher, and finally audits the edited policy before merging the accepted parameter update.}
    \Description{Diagram showing the CARVE framework workflow, from input policy and data to edited policy, with steps for attribution, selection, replacement, and auditing.}
    \label{fig:carve-design}
\end{figure*}

The strategy unlearning MARL framework \CARVE{} takes as input the original policy \(\pi_0\), its parameters \(\theta_0\), rollout logs or simulator access, the semantic request \(\varphi_f\), and optional retain specifications of behaviors or task performance to retain. It outputs a merged policy \(\pi^-\) whose actor parameters are changed. The automaton is used during unlearning and auditing, but not as an inference-time shield. Figure~\ref{fig:carve-design} summarizes the four stages.

\subsection{Causal Coordination Attribution}

The first stage asks a simple question: which agents, at which times, make the forbidden convention more likely? We estimate two centralized critics on the product state \(\bar s_t=(s_t,q_t)\):
\begin{equation}
Q^J(\bar s_t,\mathbf a_t)=
\mathbb E\!\left[
\sum_{k=t}^{H-1}\gamma^{k-t}r_k
\mid
\bar s_t,\mathbf a_t
\right],
Q^\varphi(\bar s_t,\mathbf a_t)=
\Pr(q_H\in F\mid\bar s_t,\mathbf a_t).
\end{equation}
\(Q^J\) estimates future task return, and \(Q^\varphi\) estimates the probability that the monitor eventually accepts. Because \(q_t\) records formula progress, \(Q^\varphi\) can assign credit to early decisions that only become visibly forbidden later.

To measure an agent's contribution, we compare the observed joint action with counterfactual actions sampled from a task-plausible baseline \(b\). For a coalition \(S\subseteq\mathcal N\),
\begin{equation}
v_t(S)=
\mathbb E_{\mathbf a'_{\bar S,t}\sim b}
\left[
Q^\varphi(\bar s_t,\mathbf a_{S,t},\mathbf a'_{\bar S,t})
\right],
\end{equation}
where $\mathbf a_{S,t}$ keeps the observed actions of agents in $S$, and $\mathbf a'_{\bar S,t}$ replaces the other agents' current actions. Specifically, $b$ samples alternative actions from the original policy's high-probability actions at the current local histories, excluding the observed actions and renormalizing their probabilities. A Shapley-style attribution is then
\begin{equation}
\label{eq:individual-attribution}
\chi_{i,t}=
\mathbb E_{S\sim\omega_i}
\left[
v_t(S\cup\{i\})-v_t(S)
\right],
\end{equation}
where \(\omega_i\) is the standard Shapley distribution~\cite{shapley1953value,you2026joint} over coalitions not containing \(i\). Thus \(\chi_{i,t}>0\) means that restoring agent \(i\)'s original action increases semantic risk on average. To keep pairwise coordination visible~\cite{grabisch1999axiomatic}, we also estimate
\begin{equation}
\label{eq:pairwise-attribution}
\chi_{ij,t}^{(2)}=
\mathbb E_{S\sim\omega_{ij}}
\left[
v_t(S\cup\{i,j\})
-v_t(S\cup\{i\})
-v_t(S\cup\{j\})
+v_t(S)
\right].
\end{equation}
This second score is positive when two agents jointly create more risk than their separate effects explain. In practice, we can approximate these expectations with sampled coalitions rather than enumerating all subsets.

\subsection{Retain-Aware Strategy Subspace}

The second stage turns attribution into parameter directions. Let
\begin{equation}
\psi_{i,t}
=
\nabla_\theta\log\pi_{\theta_i}(a_t^i\mid h_t^i)
\end{equation}
be the policy score of agent \(i\)'s decision. We convert the attribution scores into one nonnegative weight
\begin{equation}
\label{eq:attribution-weight}
w_{i,t}=
[\chi_{i,t}]_+
+
\eta\sum_{j\neq i}[\chi_{ij,t}^{(2)}]_+,
\end{equation}
where \([x]_+=\max(x,0)\) and \(\eta\) controls how much pairwise coordination affects the weight. The forbidden-strategy direction for trajectory \(\tau\) is
\begin{equation}
\label{eq:forbidden-direction}
g_f(\tau)=
\sum_{i,t}w_{i,t}\psi_{i,t}.
\end{equation}
This vector points toward parameters that support decisions causally involved in \(\varphi_f\).

Aggregating over high-risk trajectories gives
\begin{equation}
G_f=\mathbb E[g_f(\tau)g_f(\tau)^\top],
\end{equation}
the forbidden-strategy sensitivity matrix. 

We also estimate a retained Fisher matrix~\citep{yang2022efficient} \(F_r\) from trajectories and decision points that do not trigger the target convention and should remain stable. Its empirical estimate is computed as
\begin{equation}
\widehat F_r=\frac{1}{M}\sum_{(i,t)\in\mathcal I_r}\psi_{i,t}\psi_{i,t}^{\top},
\end{equation}
where \(\mathcal I_r\) is the set of retained decision points, \(M=|\mathcal I_r|\), and all policy score vectors are evaluated at the original parameters \(\theta_0\). Intuitively, \(u^\top G_f u\) is large if direction \(u\) changes the forbidden strategy, while \(u^\top F_r u\) is large if \(u\) disturbs retained behavior.

\CARVE{} chooses edit directions by solving
\begin{equation}
G_f u_k = \lambda_k(F_r+\lambda_0 I)u_k.
\label{eq:retain-aware-eigenproblem}
\end{equation}
Here \(u_k\) is an edit direction, \(\lambda_k\) is its strategy-to-retain ratio, and \(\lambda_0>0\) stabilizes the retain metric. Let $B=[u_1,\ldots,u_K]$ contain the top-$K$ candidate directions. To obtain sparse edits, we select a fixed coordinate mask $m_f\in\{0,1\}^{P}$ containing the top $\lceil fP\rceil$ coordinates under the retain-aware semantic sensitivity score,
and form \(B_f=\operatorname{diag}(m_f)B\), which is a sparsification heuristic.
Subsequent optimization is restricted to
\begin{equation}
\theta_\alpha=\theta_0+B_f\alpha,
\qquad \alpha\in\mathbb R^K.
\label{eq:subspace-update}
\end{equation}
Only $\alpha$ is optimized; the masked basis remains fixed.


\subsection{Counterfactual Re-Coordination}

Removing a convention can leave a coordination gap: agents may stop using the forbidden strategy but fail to coordinate on an alternative. This stage therefore builds a counterfactual teacher at product states with high \(Q^\varphi\). Here, counterfactual refers to alternative joint actions evaluated at the same decision context to construct replacement coordination. For each such state, the teacher favors joint actions that keep task value high and semantic risk low:
\begin{equation}
\label{eq:cf-teacher}
\widetilde\mu_t(\mathbf a\mid\bar s_t)
=
\frac{1}{Z_t}
\boldsymbol\pi_0(\mathbf a\mid\mathbf h_t)
\exp\!\left(
\beta Q^J(\bar s_t,\mathbf a)
-
\lambda Q^\varphi(\bar s_t,\mathbf a)
\right).
\end{equation}
\(Z_t\) normalizes the distribution over joint actions. The coefficient \(\beta\ge0\) rewards useful alternatives, and \(\lambda\ge0\) penalizes actions likely to complete the forgotten convention. This is a compact way to express a KL-near replacement teacher: stay close to \(\pi_0\), but tilt probability mass away from the target strategy and toward valuable alternatives. Because the teacher is centralized, we distill its agent-wise marginals \(\widetilde\mu_{i,t}\) into the decentralized actors:
\begin{equation}
\label{eq:cf-loss}
\mathcal L_{\mathrm{cf}}(\alpha)=
\mathbb E_{\widetilde\tau}
\sum_{i,t}w_{i,t}
D_{\mathrm{KL}}\!\left(
\widetilde\mu_{i,t}(\cdot\mid\bar s_t)
\Vert
\pi_{\theta_\alpha,i}(\cdot\mid h_t^i)
\right).
\end{equation}
\(\widetilde\tau\) denotes teacher-generated or teacher-relabeled trajectories, and \(\widetilde\mu_{i,t}\) is the teacher's marginal distribution for agent \(i\). The same attribution weight \(w_{i,t}\) focuses distillation on decisions that matter for the target convention, rather than forcing the entire policy to imitate the teacher everywhere.

\subsection{Robust Auditing and Policy Merge}

The final stage chooses \(\alpha\) and repeatedly audits the candidate policy \(\pi_{\theta_\alpha}\). The training objective contains only two terms:
\begin{equation}
\label{eq:projection-objective}
\min_{\alpha}\quad\mathcal L_{\mathrm{cf}}(\alpha)+\lambda_{\mathrm{r}}\mathcal L_{\mathrm{retain}}(\alpha),
\end{equation}
where
\begin{equation}
\label{eq:retain-loss}
\mathcal L_{\mathrm{retain}}(\alpha)
=
\mathbb E_{\tau\sim\mathcal D_r}
\sum_{i,t}
D_{\mathrm{KL}}\!\left(
\pi_{0,i}(\cdot\mid h_t^i)
\Vert
\pi_{\theta_\alpha,i}(\cdot\mid h_t^i)
\right)
\end{equation}
keeps target-unrelated behavior close to the original policy. The semantic monitor is used for auditing:
\begin{equation}
\label{eq:adversarial-risk}
\widehat{\Rphi}_{\mathrm{adv}}(\pi)
=
\max_{c\in\mathcal C_{\mathrm{adv}}}
\widehat{\Rphi}(\pi;c)
\end{equation}
is the empirical worst-case semantic risk over adversarial contexts. These contexts include held-out partners, starts, layouts, and role permutations selected to elicit \(\varphi_f\). If the audit risk is too high, the eliciting trajectories are added to the training set and the update is repeated. 
We accept the final policy only if it passes the semantic-risk and return checks from the formulation section. The delivered model is obtained by merging \(\Delta\theta=B_f\alpha\) into the actor parameters; no monitor or shield is required at deployment.

\section{Experiments}
We conduct experiments to answer four questions about \CARVE{}: \textbf{i). RQ1}: whether it removes the requested coordination strategy (\S\ref{sec:q1.0}, \S\ref{sec:q1.2}, \S\ref{sec:vis})? \textbf{ii). RQ2}: whether it preserves MARL task performance (\S\ref{sec:q2})? \textbf{iii). RQ3}: whether it is efficient enough to be a post-training edit rather than retraining in disguise (\S\ref{sec:q3})? \textbf{iv). RQ4}: which components and hyperparameters control the erasure–retention frontier (\S\ref{sec:q4},\S\ref{sec:para})?

\subsection{Experimental Setup}
\subsubsection{Testbed}
Experiments are implemented in Python 3.11.8 with PyTorch 2.6.0+cu124 and TorchRL 0.7.2. They run on an Ubuntu 18.04.6 LTS server with two 26-core Intel Xeon Platinum 8269CY CPUs, 256 GB RAM, and four NVIDIA A100-40GB GPUs. Unless otherwise stated, backbone training uses five random seeds.

\subsubsection{Evaluation Benchmarks and Unlearning Requests}
Our evaluation cross five MARL environments pairs established cooperative MARL testbeds with role-invariant unlearning requests. For SMACv2 \citep{ellis2023smacv2}, the target is a relational combat tactic; for Overcooked-AI/OGC\citep{carroll2019utility,ruhdorfer2024overcooked}, it is a bottleneck or handoff convention; for VMAS\citep{bettini2022vmas}, it is a continuous-control yielding or leader-follower rule; for RWARE \citep{papoudakis2020benchmarking}, it is a warehouse aisle convention; and for LBF\citep{papoudakis2020benchmarking}, it is a sparse-reward partner-selection convention, which is used as an additional sparse-reward stress test. Details of the scenarios, role-invariant unlearning requests, retain specifications, backbone algorithms, and hyperparameters, are provided in Appendix~\ref{sec:a_D_Details}.

\subsubsection{Baselines}
The baselines are chosen to answer distinct alternative explanations: whether runtime shielding merely hides the behavior, whether constrained MARL is enough, whether generic parameter unlearning is enough, and whether trajectory removal is enough. We compare against the methods as shown in Table \ref{tab:baselines}.
\begin{table}[t!]
\centering
\setlength{\abovecaptionskip}{0.2cm}
\setlength{\tabcolsep}{3.5pt}
\caption{Baseline suite used in the main result tables.}
\renewcommand{\arraystretch}{1.05}
\begin{tabular}{@{}p{0.25\linewidth}p{0.75\linewidth}@{}}
\toprule
Method & Role in evaluation \\
\midrule
Orig. (trained backbone) & Pre-unlearning policy and target-strategy prevalence. \\
\midrule
Shield (runtime automaton) & Blocks forbidden completions without model updates. \\
\CARVE{} (Ours) & Sparse retain-aware semantic strategy unlearning. \\
Full-FT & Full-parameter semantic fine-tuning without sparse subspace using negative strategy gradient. \\
RP (semantic reward penalty) & Fixed semantic-cost penalty during post-training fine-tuning. \\
Lag. (Lagrangian MARL)~\citep{gu2021multi} & Adaptive Lagrangian penalty for semantic-cost suppression. \\
MACPO~\citep{gu2021multi} & Conservative constrained policy optimization baseline. \\
Fisher~\citep{foster2024fast} & Fisher-ratio parameter dampening without causal attribution. \\
TrajDel~\citep{gong2024trajdeleter} & Tests trajectory removal for role-invariant strategies. \\
\bottomrule
\end{tabular}
\label{tab:baselines}
\end{table}

\begin{table}[t!]
\centering
\caption{Unlearning performance across five MARL benchmarks.
OC-AI denotes Overcooked-AI.}
\label{tab:unlearning-performance}
\normalsize
\setlength{\tabcolsep}{2pt}
\renewcommand{\arraystretch}{1.0}
\begin{tabular}{@{}lc|c|c|c|c|r@{}}
\toprule
Method & Metric
& \shortstack{SMACv2\\Zerg}
& \shortstack{OC-AI\\Asym.}
& \shortstack{VMAS\\Flocking}
& \shortstack{RWARE\\Tiny}
& \shortstack{LBF\\8x8-2p} \\
\hline
\multirow{2}{*}{Orig.}
& SSR$\downarrow$ & 0.469 & 1.000 & 0.357 & 0.875 & 0.792 \\
& WER$\downarrow$ & 0.875 & 1.000 & 0.793 & 0.875 & 0.792 \\
\hline
\multirow{2}{*}{Shield}
& SSR$\downarrow$ & 0.125 & 0.000 & 0.180 & 0.000 & 0.000 \\
& WER$\downarrow$ & 0.500 & 0.000 & 0.480 & 0.000 & 0.000 \\
\multirow{2}{*}{\CARVE{}}
& SSR$\downarrow$ & 0.144 & 0.000 & 0.157 & 0.000 & 0.000 \\
& WER$\downarrow$ & 0.500 & 0.000 & 0.222 & 0.000 & 0.000 \\
\multirow{2}{*}{Full-FT}
& SSR$\downarrow$ & 0.312 & 0.000 & 0.292 & 0.000 & 0.000 \\
& WER$\downarrow$ & 0.875 & 0.000 & 0.361 & 0.000 & 0.000 \\
\multirow{2}{*}{RP}
& SSR$\downarrow$ & 0.250 & 0.000 & 0.187 & 0.000 & 0.000 \\
& WER$\downarrow$ & 0.750 & 0.000 & 0.274 & 0.000 & 0.000 \\
\multirow{2}{*}{Lag.}
& SSR$\downarrow$ & 0.344 & 0.000 & 0.219 & 0.000 & 0.000 \\
& WER$\downarrow$ & 0.625 & 0.000 & 0.505 & 0.000 & 0.000 \\
\multirow{2}{*}{MACPO}
& SSR$\downarrow$ & 0.203 & 0.000 & 0.164 & 0.000 & 0.000 \\
& WER$\downarrow$ & 0.500 & 0.000 & 0.369 & 0.000 & 0.000 \\
\multirow{2}{*}{Fisher}
& SSR$\downarrow$ & 0.422 & 0.000 & 0.399 & 0.000 & 0.042 \\
& WER$\downarrow$ & 0.750 & 0.000 & 0.743 & 0.000 & 0.042 \\
\addlinespace[1pt]
\multirow{2}{*}{TrajDel}
& SSR$\downarrow$ & 0.403 & 1.000 & 0.292 & 0.750 & 0.792 \\
& WER$\downarrow$ & 0.625 & 1.000 & 0.586 & 0.750 & 0.792 \\
\bottomrule
\end{tabular}
\end{table}

\subsubsection{Evaluation Metrics}
We assess convention removal using the nominal semantic satisfaction rate $\mathrm{SSR}(\pi)=\Pr_{\tau\sim\pi}(\tau\models\varphi_f)$
and the worst-case elicitation rate $\mathrm{WER}(\pi)=
\max_{\nu\in\mathcal V}\Pr_{\pi,\nu}(\tau\models\varphi_f).$
Here, \(\mathcal V\) denotes the red-team elicitation set, including held-out teammates or partners when applicable, held-out starts, layouts, procedural context seeds, and role permutations. Both metrics measure the occurrence of the forbidden convention, with lower values indicating more effective suppression. We assess task retention using episodic return (Return) and normalized task-utility retention ($\mathrm{Util.}(\pi)=\operatorname{clip}_{[0,1]}\!\left(1-\frac{[J_0-J(\pi)]_+}{D_b}\right)$), where \(J_0\) is the original policy's mean return recorded in the same run, \([x]_+=\max(x,0)\). The evaluation scripts use \(D_b=\max(|J_0|,1)\) for SMACv2 and VMAS, which accommodates negative and near-zero original returns, and \(D_b=J_0\) for Overcooked-AI, RWARE, and LBF, whose reported original returns are positive. Util. is normalized relative to the restored original policy, with \(1\) indicating full task-performance retention. We measure efficiency using unlearning speedup \(T_{\mathrm{backbone}}/T_{\mathrm{unlearn}}\), inference latency (Infer.), and the fraction of actor parameters changed relative to the original checkpoint (ParamFrac).

\begin{table}[t!]
\centering
\caption{MARL retain performance across five benchmarks.
OC-AI denotes Overcooked-AI.}
\label{tab:retain-performance}
\normalsize
\setlength{\tabcolsep}{2pt}
\renewcommand{\arraystretch}{1.0}
\begin{tabular}{@{}lc|c|c|c|c|r@{}}
\toprule
Method & Metric
& \shortstack{SMACv2\\Zerg}
& \shortstack{OC-AI\\Asym.}
& \shortstack{VMAS\\Flocking}
& \shortstack{RWARE\\Tiny}
& \shortstack{LBF\\8x8-2p} \\
\hline
\multirow{2}{*}{Orig.}
& Return & 8.407 & 216.0 & 0.483 & 1.000 & 0.667 \\
& Util.  & 1.000 & 1.000 & 1.000 & 1.000 & 1.000 \\
\hline
\multirow{2}{*}{Shield}
& Return & 4.025 & 0.0 & 0.481 & 0.000 & 0.000 \\
& Util.  & 0.749 & 0.000 & 0.902 & 0.000 & 0.000 \\
\multirow{2}{*}{\CARVE{}}
& Return & 7.268 & 109.0 & 0.557 & 1.250 & 0.625 \\
& Util.  & 0.941 & 0.542 & 1.000 & 1.000 & 0.938 \\
\multirow{2}{*}{Full-FT}
& Return & 5.111 & 0.0 & 0.401 & 0.000 & 0.000 \\
& Util.  & 0.802 & 0.000 & 0.918 & 0.000 & 0.000 \\
\multirow{2}{*}{RP}
& Return & 4.163 & 0.0 & 0.426 & 0.000 & 0.000 \\
& Util.  & 0.673 & 0.000 & 0.943 & 0.000 & 0.000 \\
\multirow{2}{*}{Lag.}
& Return & 4.702 & 0.0 & 0.405 & 0.000 & 0.000 \\
& Util.  & 0.747 & 0.000 & 0.922 & 0.000 & 0.000 \\
\multirow{2}{*}{MACPO}
& Return & 4.524 & 0.0 & 0.460 & 0.000 & 0.000 \\
& Util.  & 0.544 & 0.000 & 0.976 & 0.000 & 0.000 \\
\multirow{2}{*}{Fisher}
& Return & 7.211 & 0.0 & 0.526 & 0.000 & 0.021 \\
& Util.  & 0.903 & 0.000 & 1.000 & 0.000 & 0.031 \\
\multirow{2}{*}{TrajDel}
& Return & 4.783 & 207.0 & 0.400 & 0.875 & 0.667 \\
& Util.  & 0.803 & 0.958 & 0.916 & 0.875 & 1.000 \\
\bottomrule
\end{tabular}
\end{table}

\subsection{Overall Performance}

\subsubsection{Unlearning Performance}
\label{sec:q1.0}
Table~\ref{tab:unlearning-performance} shows that \CARVE{} reduces mean SSR and WER by 91.4\% and 83.3\%, respectively, relative to the original policies. No target-strategy completions are observed under either protocol in Overcooked-AI, RWARE, and LBF. In the two benchmarks with residual risk, \CARVE{} matches the lowest reported WER in SMACv2 and achieves the lowest WER in VMAS. Relative to MACPO, the strongest parameter-editing baseline by mean WER, the largest within-benchmark SSR and WER reductions reach 29.1\% and 39.8\%, respectively. Compared with Shield, \CARVE{} matches WER in four benchmarks and reduces it by 53.8\% in VMAS, while requiring no inference-time monitor. For Full-FT, RP, and Lag., the largest within-benchmark WER reductions range from 33.3\% to 56.0\%. These gains are consistent with targeting causally attributed coordination decisions, constructing useful alternatives, and incorporating eliciting trajectories during auditing, rather than relying solely on broad updates or semantic penalties. Fisher leaves greater residual semantic risk, while TrajDel still permits target-strategy completions in every benchmark. This shows that parameter dampening may miss the interactions underlying a convention, while deleting selected trajectories does not prevent the strategy's reproduction through other histories. SMACv2 remains challenging, with a residual WER of \(0.500\). This matches the lowest observed WER among the evaluated methods. At this risk level, however, \CARVE{} retains a utility of \(0.941\), compared with \(0.749\) for Shield and \(0.544\) for MACPO. Thus, although substantial residual risk remains, \CARVE{} preserves the most task utility among methods achieving the same worst-case suppression. These results highlight the importance of evaluating convention removal jointly with retained coordination competence.

\subsubsection{MARL Retain Performance}
\label{sec:q2}
Table~\ref{tab:retain-performance} shows \CARVE{} reduces a median utility of only 5.9\% relative to the original policies, with four of the five benchmarks losing at most 6.2\%. The retention comparison is particularly informative in Overcooked-AI, RWARE, and LBF: Shield, Full-FT, RP, Lag., and MACPO achieve zero observed SSR/WER alongside zero task return. Their suppression is therefore accompanied by task failure, whereas \CARVE{} maintains positive returns in all three settings. In SMACv2, \CARVE{} matches Shield and MACPO in WER while improving return by up to 80.6\%. This pattern is consistent with the intended combination of retain-aware subspace selection, KL retention, and teacher-guided replacement coordination. TrajDel preserves slightly higher mean utility, but the target convention remains elicitable on every benchmark. Relative to TrajDel, \CARVE{} gives up 2.9\% in mean utility while reducing mean WER by 80.8\%. Fisher preserves utility well in SMACv2 and VMAS but does not consistently retain performance across the remaining tasks. Overcooked-AI exhibits the largest utility loss with a \(45.8\%\) reduction. Productive workflows can depend strongly on the convention being removed, making effective replacement coordination difficult. Nevertheless, among methods attaining zero observed SSR and WER, \CARVE{} achieves the highest retained utility; all corresponding baselines yield zero task return. These comparisons show that \CARVE{} offers the strongest observed retention at the lowest reported semantic risk in this setting. The remaining utility gap may reflect both dependence on the removed convention and limitations of replacement coordination.

\begin{table}[t!]
\centering
\setlength{\abovecaptionskip}{0.2cm}
\caption{Stronger unlearning audit under two different attacks on three VMAS forgetting tasks. Bold numbers indicate the largest reported value among the three audit settings.}
\renewcommand{\arraystretch}{1}
\resizebox{\linewidth}{!}{
\begin{tabular}{@{}c|l|ccc|ccc@{}}
\toprule
Task & Method & SSR$\downarrow$ & Return$\uparrow$ & Util.$\uparrow$ & Norm. WER & Role WER & Ctx. WER \\
\midrule
\multirow{9}{*}{\rotatebox[origin=c]{90}{\textbf{GiveWay (MADDPG)}}}
 & Orig. & 0.502 & 3.708 & 1.000 & 0.502 & 0.509 & \textbf{0.517} \\
 \cline{2-8}
 & Shield & 0.326 & 3.125 & 0.982 & 0.326 & 0.399 & \textbf{0.401} \\
 & \CARVE{} & 0.313 & 3.679 & 0.998 & 0.313 & 0.351 & \textbf{0.360} \\
 & Full-FT & 0.399 & 3.172 & 0.990 & 0.399 & 0.407 & \textbf{0.432} \\
 & RP & 0.404 & 3.276 & 0.991 & 0.404 & 0.416 & \textbf{0.423} \\
 & Lag. & 0.425 & 3.259 & 0.987 & 0.425 & 0.398 & \textbf{0.431} \\
 & MACPO & 0.405 & 3.170 & 0.990 & \textbf{0.405} & 0.374 & 0.395 \\
 & Fisher & 0.481 & 3.152 & 0.985 & 0.481 & 0.487 & \textbf{0.491} \\
 & TrajDel & 0.501 & 2.713 & 0.892 & 0.501 & 0.501 & \textbf{0.506} \\
\midrule
\multirow{9}{*}{\rotatebox[origin=c]{90}{\textbf{Passage (MAPPO)}}}
 & Orig. & 0.338 & 47.5 & 1.000 & 0.755 & 0.747 & \textbf{0.756} \\
 \cline{2-8}
 & Shield & 0.203 & 29.6 & 0.623 & 0.341 & 0.360 & \textbf{0.374} \\
 & \CARVE{} & 0.208 & 158.2 & 1.000 & 0.340 & \textbf{0.351} & 0.343 \\
 & Full-FT & 0.248 & 103.1 & 1.000 & 0.365 & \textbf{0.398} & 0.367 \\
 & RP & 0.217 & 128.1 & 1.000 & 0.373 & \textbf{0.406} & 0.362 \\
 & Lag. & 0.229 & 133.9 & 1.000 & 0.512 & 0.521 & \textbf{0.623} \\
 & MACPO & 0.213 & 143.5 & 1.000 & 0.406 & 0.387 & \textbf{0.441} \\
 & Fisher & 0.441 & -103.7 & 0.000 & 0.821 & \textbf{0.853} & 0.836 \\
 & TrajDel & 0.337 & 48.5 & 1.000 & \textbf{0.754} & 0.745 & \textbf{0.754} \\
\midrule
\multirow{9}{*}{\rotatebox[origin=c]{90}{\textbf{Flocking (IPPO)}}}
 & Orig. & 0.357 & 0.483 & 1.000 & 0.793 & \textbf{0.797} & 0.797 \\
 \cline{2-8}
 & Shield & 0.180 & 0.481 & 0.902 & 0.480 & 0.501 & \textbf{0.541} \\
 & \CARVE{} & 0.157 & 0.557 & 1.000 & 0.222 & \textbf{0.290} & 0.280 \\
 & Full-FT & 0.292 & 0.401 & 0.918 & 0.361 & \textbf{0.443} & 0.387 \\
 & RP & 0.187 & 0.426 & 0.943 & 0.274 & \textbf{0.358} & 0.271 \\
 & Lag. & 0.219 & 0.405 & 0.922 & 0.505 & \textbf{0.653} & 0.533 \\
 & MACPO & 0.164 & 0.460 & 0.976 & 0.369 & \textbf{0.546} & 0.435 \\
 & Fisher & 0.399 & 0.526 & 1.000 & 0.743 & \textbf{0.858} & 0.813 \\
 & TrajDel & 0.292 & 0.400 & 0.916 & 0.586 & \textbf{0.737} & 0.719 \\
\bottomrule
\end{tabular}
}
\label{tab:strong-attack-unlearning}
\end{table}

\subsubsection{Robustness with Stronger Elicitation under Different Unlearning Requests}
\label{sec:q1.2}
We further test whether the unlearning performance persist under stronger elicitation across the GiveWay, Passage, and Flocking requests in VMAS dataset, using their respective MADDPG~\citep{lowe2017multi}, MAPPO~\citep{yu2022surprising}, and IPPO~\citep{de2020independent} checkpoints. We use two red-team attacks. The role-permutation attack swaps agent identities before policy evaluation and maps the resulting actions back before stepping. The role-invariant semantic request remains unchanged, so this attack tests whether reassignment of the same roles to different agent indices can recover the target convention. The held-out-context attack evaluates the same edited checkpoint on unseen VMAS context seeds \(\{4001,4002,4003\}\), which vary initial states and procedural task contexts. No additional parameter updates are performed during these tests; \CARVE{} executes its merged actors, and the semantic monitor scores the resulting trajectories.

Table~\ref{tab:strong-attack-unlearning} reports standard-context WER (Norm. WER), WER under role permutation (Role WER), and the maximum WER across held-out-context seeds (Ctx. WER). Under role permutation, \CARVE{} achieves the lowest reported Role WER for all three unlearning requests. Relative to the original policies, the reduction across requests has a median of 53.0\% and reaches 63.6\%. Across parameter-editing baselines, the median Role WER reductions range from 9.3\% to 58.9\%, with the largest reduction against MACPO reaching 46.9\%. Compared with Shield, the median reduction is 12.0\%, and the largest reaches 42.1\%. These comparisons show that \CARVE{}'s advantage persists when the same roles are reassigned to different agent indices. Under held-out contexts, \CARVE{} reduces Ctx. WER relative to the original policies by a median of 54.6\%, with the largest reduction reaching 64.9\%. Its median Ctx. WER reductions against parameter-editing baselines range from 5.2\% to 59.0\%, including 22.2\% against MACPO. Relative to Shield, the median reduction is 10.2\%, and the largest reaches 48.2\%. \CARVE{} achieves the lowest Ctx. WER in GiveWay and Passage, while RP has slightly lower Ctx. WER in Flocking. When risk is summarized by the maximum over the nominal, role-permutation, and held-out-context protocols for each method and request, \CARVE{} nevertheless achieves the lowest value for all three requests. This supports its comparative robustness across the tested role assignments and unseen contexts, while residual strategy completions remain possible.

\begin{table*}[t!]
\centering
\setlength{\abovecaptionskip}{0.2cm}
\caption{Unlearning efficiency across five MARL benchmarks.
Speedup is $T_{\mathrm{backbone}}/T_{\mathrm{unlearn}}$;
Infer. is inference latency.}
\resizebox{0.8\linewidth}{!}{
\begin{tabular}{@{}l|l|r|rrrrrrrr@{}}
\toprule
Benchmark & Metric & Orig. & Shield & \CARVE{} & Full-FT & RP & Lag. & MACPO & Fisher & TrajDel \\
\hline
\multirow{2}{*}{SMACv2 (Zerg)}
 & Speedup($\times$)$\uparrow$ & -- & -- & 584.83 & 255.06 & 290.30 & 266.87 & 273.08 & 575.38 & 482.07 \\
 & Infer.(ms)$\downarrow$ & 120.326 & 143.460 & 120.326 & 120.326 & 120.326 & 120.326 & 120.326 & 120.326 & 120.326 \\
\hline
\multirow{2}{*}{Overcooked-AI (Asym.)}
 & Speedup($\times$)$\uparrow$ & -- & -- & 101.92 & 44.77 & 48.68 & 45.59 & 47.20 & 94.32 & 82.37 \\
 & Infer.(ms)$\downarrow$ & 80.750 & 104.776 & 80.750 & 80.750 & 80.750 & 80.750 & 80.750 & 80.750 & 80.750 \\
\hline
\multirow{2}{*}{VMAS (Flocking)}
 & Speedup($\times$)$\uparrow$ & -- & -- & 454.53 & 363.68 & 391.25 & 326.47 & 357.24 & 416.54 & 423.68 \\
 & Infer.(ms)$\downarrow$ & 86.718 & 107.856 & 86.718 & 86.718 & 86.718 & 86.718 & 86.718 & 86.718 & 86.718 \\
\hline
\multirow{2}{*}{RWARE (Tiny)}
 & Speedup($\times$)$\uparrow$ & -- & -- & 247.47 & 178.54 & 199.10 & 182.61 & 188.69 & 228.85 & 209.02 \\
 & Infer.(ms)$\downarrow$ & 100.735 & 122.740 & 100.735 & 100.735 & 100.735 & 100.735 & 100.735 & 100.735 & 100.735 \\
\hline
\multirow{2}{*}{LBF (8x8-2p)}
 & Speedup($\times$)$\uparrow$ & -- & -- & 312.94 & 209.17 & 228.46 & 216.34 & 220.46 & 264.57 & 291.89 \\
 & Infer.(ms)$\downarrow$ & 70.605 & 92.617 & 70.605 & 70.605 & 70.605 & 70.605 & 70.605 & 70.605 & 70.605 \\
\bottomrule
\end{tabular}
}
\label{tab:unlearning-efficiency}
\end{table*}

\subsection{Unlearning Efficiency}
\label{sec:q3}
\label{subsec:unlearning-efficiency}
\begin{figure}[t!]
    \centering
    \includegraphics[width=0.8\linewidth]{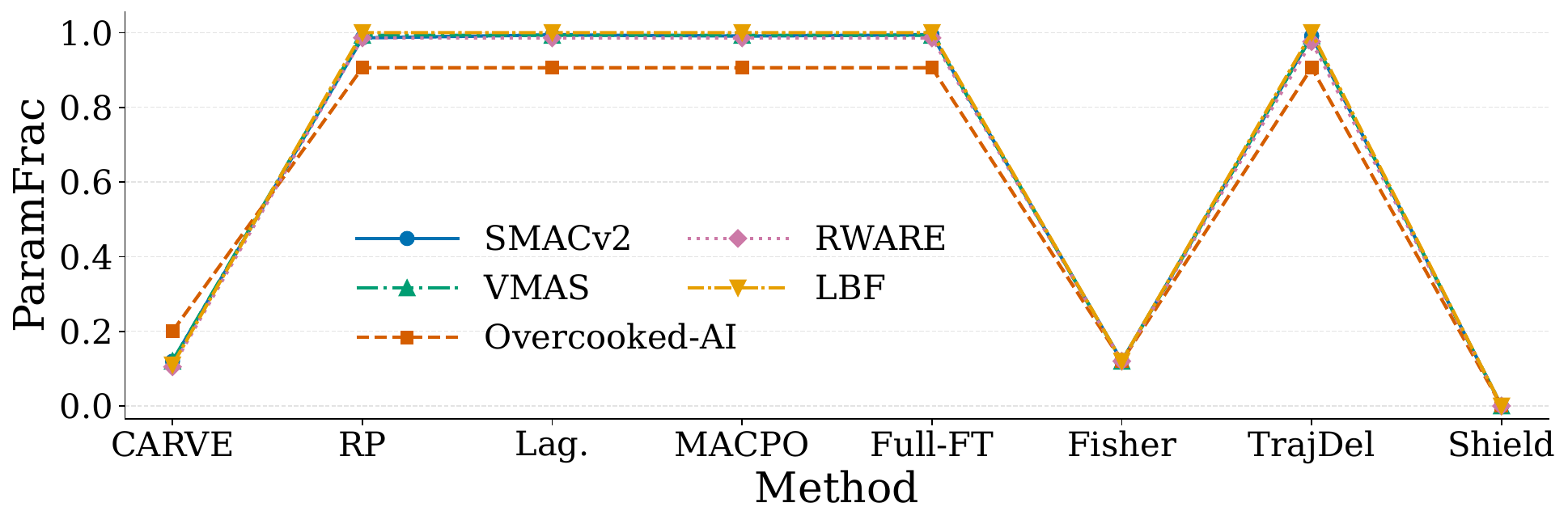}
    \caption{Fraction of actor parameters modified by each unlearning method across five MARL benchmarks}
    \Description{Bar chart showing the fraction of actor parameters modified by each unlearning method across five MARL benchmarks.}
    \label{fig:paramfrac}
\end{figure}

Table~\ref{tab:unlearning-efficiency} shows that \CARVE{} achieves the highest reported unlearning time speedup across all five benchmarks, ranging from \(101.92\times\) to \(584.83\times\) relative to original backbone training, including loading data, gradient-mask construction, and update time. It is \(1.25\)--\(2.29\) times faster than Full-FT. Its inference latency matches the original policy, while Shield increases latency by \(19.2\%\)--\(31.2\%\). This overhead arises from additional runtime automaton monitoring and action filtering. Figure~\ref{fig:paramfrac} reports \textit{ParamFrac}, the fraction of actor parameters changed relative to the original checkpoint. \CARVE{} modifies \(13.1\%\) on average, ranging from \(10.5\%\) to \(20.0\%\)), compared with \(90.6\%\)--\(100.0\%\) for RP, Lag., MACPO, Full-FT, and TrajDel. Fisher also has a small edit footprint of \(12.0\%\) in every benchmark, but records lower unlearning speedups than \CARVE{}. This illustrates that edit locality alone does not determine runtime, which also depends on parameter selection and optimization costs. \CARVE{} optimizes only the coefficients \(\alpha\) in the retain-aware update \(\Delta\theta=B_f\alpha\). The accepted update is merged into the actors, enabling deployment without an inference-time monitor or shield.

\begin{figure}[t!]
\centering
\setlength{\tabcolsep}{3pt}
\renewcommand{\arraystretch}{0.55}
\begin{tabular}{@{}cc@{}}
\includegraphics[width=0.420\linewidth]{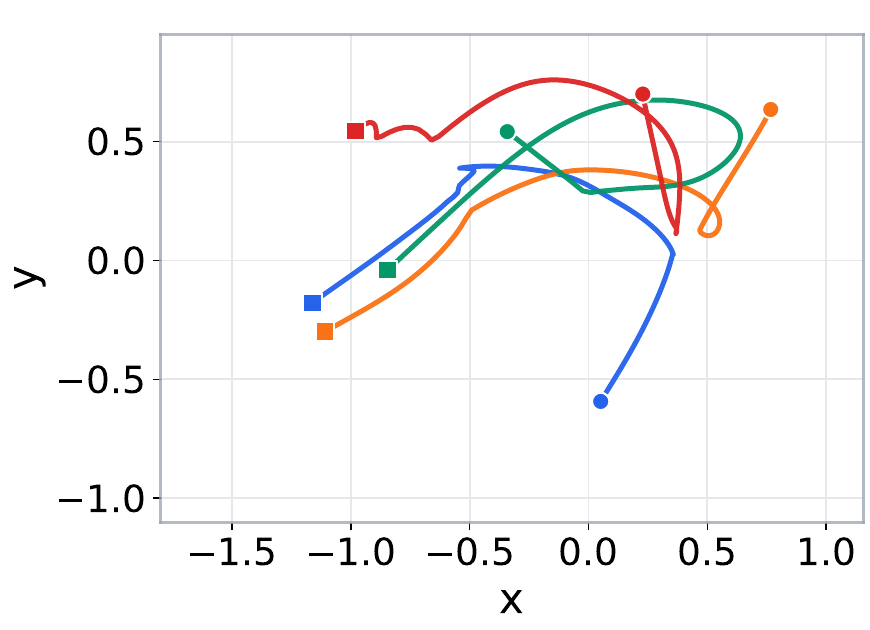} &
\includegraphics[width=0.420\linewidth]{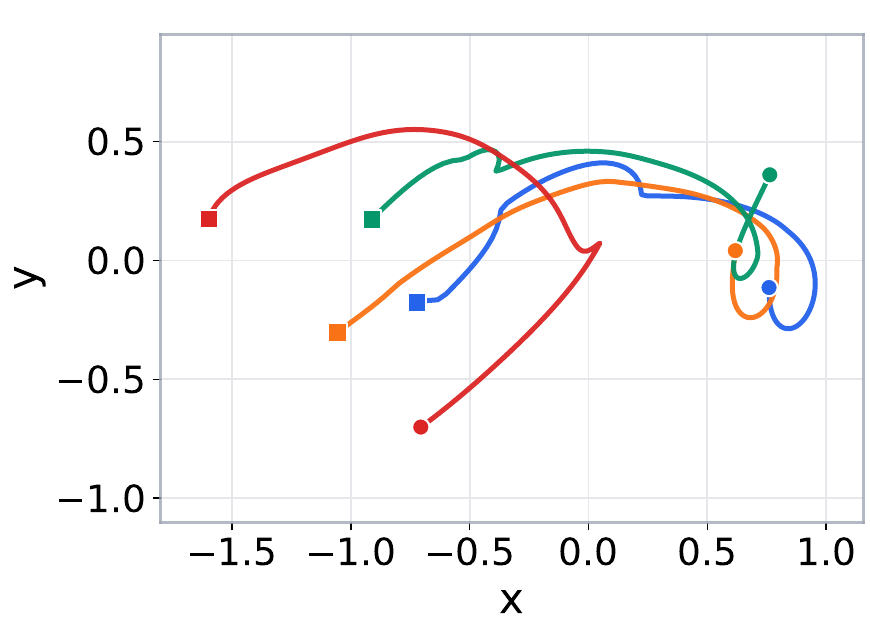} \\
\includegraphics[width=0.405\linewidth]{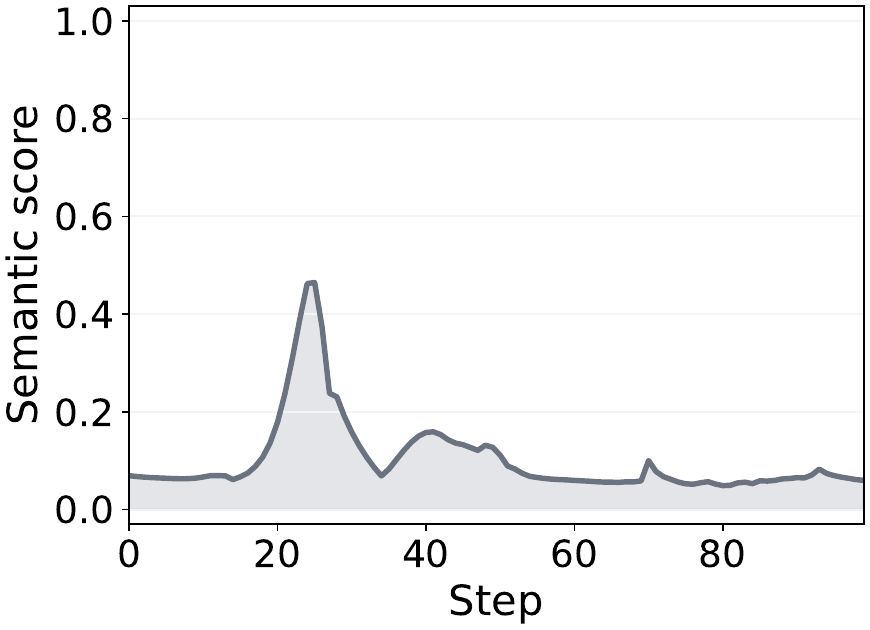} &
\includegraphics[width=0.405\linewidth]{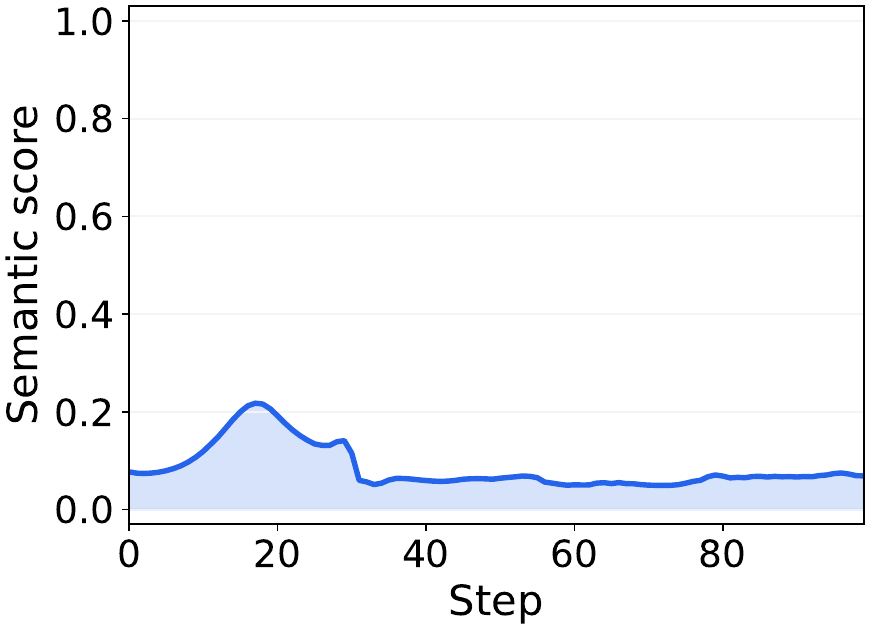} \\
\includegraphics[width=0.425\linewidth]{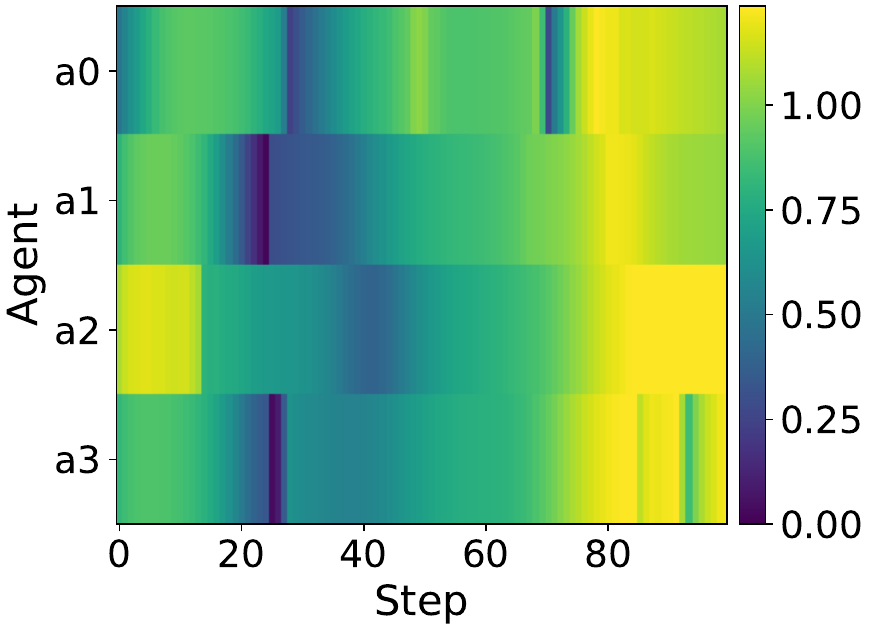} &
\includegraphics[width=0.425\linewidth]{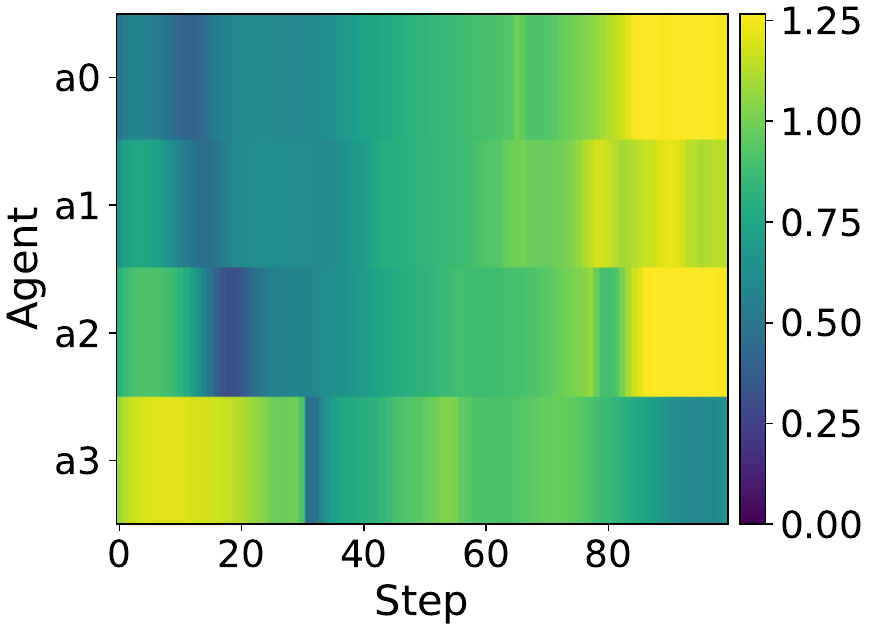} \\
\includegraphics[width=0.405\linewidth]{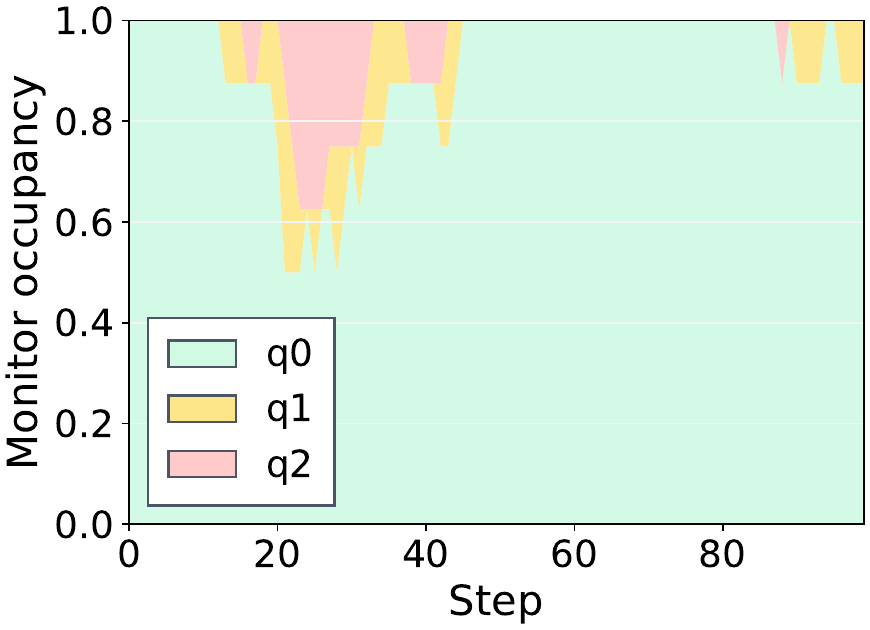} &
\includegraphics[width=0.405\linewidth]{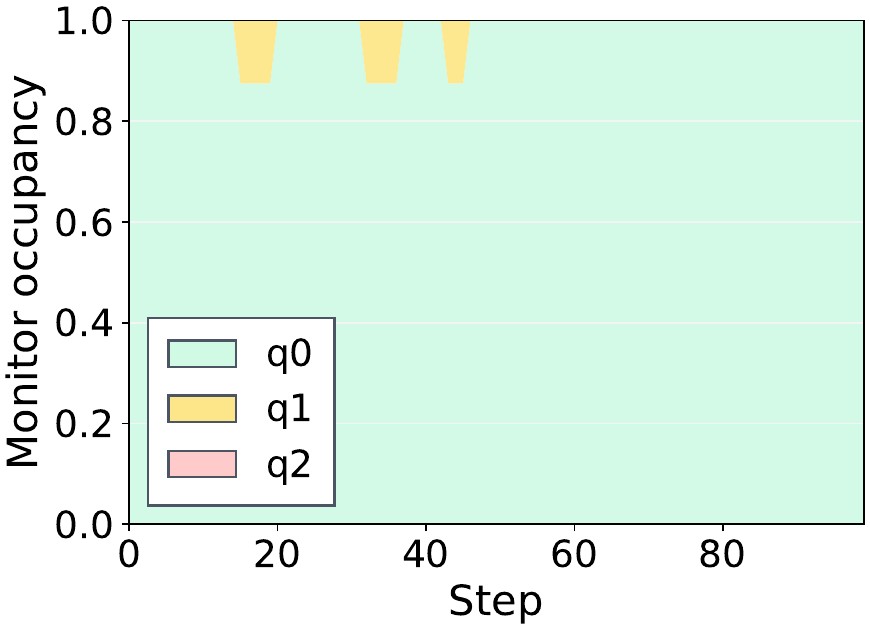}\\
(a) Original & (b) \CARVE{}
\end{tabular}
\caption{Visualization for VMAS Flocking. Rows from top to bottom show the rotated physical trajectory, the semantic score for leader-follower dominance, per-agent action norms, and occupancy of the monitor states \(q_0,q_1,q_2\).}
\label{fig:visual-flocking-panels}
\end{figure}

\subsection{Visualization}
\label{sec:vis}
We use VMAS Flocking to inspect whether semantic unlearning changes the policy itself. The target convention is leader-follower dominance: agents move cohesively while one agent exerts disproportionate control. Figure~\ref{fig:visual-flocking-panels} compares the original restored actor and \CARVE{} on the same rollout diagnostics. In the monitor row, \(q_0\), \(q_1\), and \(q_2\) denote low-risk, recoverable, and accepted forbidden-convention states. The original actor shows the target signature across all views: clustered trajectories, a semantic-score peak above the accepting threshold, uneven action norms, and visible \(q_1/q_2\) monitor occupancy. \CARVE{} weakens this signature. Its trajectories are less organized around a single leading path, the semantic score remains below the accepting region in the shown rollout, action magnitudes are more balanced, and monitor mass shifts toward \(q_0\). This supports the main claim: \CARVE{} removes the leader-follower convention while preserving coordinated flocking, rather than merely blocking actions at runtime.

\subsection{Ablation Study}
\label{sec:q4}
We ablate five components on VMAS GiveWay, Passage, and Flocking using the same checkpoints, semantic requests, retain sets, and audit contexts. The full model uses the complete pipeline in Eqs.~\eqref{eq:individual-attribution}--\eqref{eq:adversarial-risk}. The ablated variants replace the attribution scores in Eqs.~\eqref{eq:individual-attribution}--\eqref{eq:pairwise-attribution} with random sparse weights, replace the retain-aware eigenspace in Eq.~\eqref{eq:retain-aware-eigenproblem} with semantic-gradient coordinates, drop \(\mathcal L_{\mathrm{retain}}\) from Eq.~\eqref{eq:projection-objective}, drop the counterfactual teacher in Eqs.~\eqref{eq:cf-teacher}--\eqref{eq:cf-loss}, or optimize the full actor \(\theta\) instead of the subspace update in Eq.~\eqref{eq:subspace-update}. Table~\ref{tab:ablation-study} reports task averages. Causal attribution increases \(\Delta\)SSR and \(\Delta\)WER by \(147.1\%\) and \(142.0\%\), respectively, compared with random weights, supporting more effective targeting of the convention. Retain-aware selection improves \(\Delta\)WER by \(27.3\%\) and utility by \(10.5\%\), while the retention loss improves utility by \(54.9\%\) with nearly unchanged semantic reductions. These components therefore provide complementary protection for retained cooperation. Counterfactual replacement improves \(\Delta\)WER by \(43.1\%\) and mean return by \(61.6\%\), supporting productive alternative coordination. Subspace restriction improves utility by \(35.5\%\) while reducing the changed-parameter fraction by \(87.8\%\), indicating that limiting edit scope helps preserve task competence. Together, these ablations show that the components complement one another in balancing effective convention removal with efficient retained cooperation.

\begin{table}[t!]
\centering
\setlength{\abovecaptionskip}{0.2cm}
\caption{Ablation study on VMAS GiveWay, Passage, and Flocking. \(\Delta\)SSR and \(\Delta\)WER are reductions relative to the restored original checkpoint.}
\resizebox{0.9\linewidth}{!}{
\begin{tabular}{lrrrrr}
\toprule
Variant & $\Delta$SSR$\uparrow$ & $\Delta$WER$\uparrow$ & Util.$\uparrow$ & Return$\uparrow$ & ParamFrac$\downarrow$ \\
\midrule
Full \CARVE{} & 0.173 & 0.392 & 0.999 & 54.15 & 0.120 \\
\textit{w/o attribution} & 0.070 & 0.162 & 0.963 & 51.7 & 0.118 \\
\textit{w/o retain-aware mask} & 0.159 & 0.308 & 0.904 & 33.3 & 0.120 \\
\textit{w/o retain loss} & 0.174 & 0.390 & 0.645 & 32.8 & 0.120 \\
\textit{w/o replacement reg.} & 0.129 & 0.274 & 0.987 & 33.5 & 0.120 \\
\textit{w/o sparse subspace} & 0.130 & 0.340 & 0.737 & -103.6 & 0.986 \\
\bottomrule
\end{tabular}
}
\label{tab:ablation-study}
\end{table}
\nocite{1}
\nocite{li2026jeva}
\nocite{chen2026attacking}
\nocite{mao2026you}

\section{Conclusion}
This paper studied semantic coordination-strategy unlearning, where a trained cooperative policy must abandon a role-invariant convention while retaining useful cooperation. \CARVE{} combines causal coordination attribution, retain-aware subspace updates, counterfactual re-coordination, and adversarial auditing, then merges the accepted update into decentralized actors. Experiments across five MARL benchmarks show that \CARVE{} balances target-convention removal with the preservation of cooperative task performance and no additional inference latency.
\bibliographystyle{ACM-Reference-Format}
\bibliography{acmart}
\appendix
\renewcommand{\thesection}{\Alph{section}}
\renewcommand{\thesubsection}{\thesection\arabic{subsection}}

\setcounter{table}{0}
\renewcommand{\thetable}{A\arabic{table}}

\setcounter{figure}{0}
\renewcommand{\thefigure}{A\arabic{figure}}

\setcounter{equation}{0}
\renewcommand{\theequation}{A\arabic{equation}}

\ifdefined\theHtable
  \renewcommand{\theHtable}{appendix.\arabic{table}}
\fi
\ifdefined\theHfigure
  \renewcommand{\theHfigure}{appendix.\arabic{figure}}
\fi

\begin{table*}[t!]
\centering
\small
\caption{Concrete scenarios used in the main experiments. Each row reports one implemented semantic request, the corresponding retained behavior, and the single backbone used for the reported checkpoint.}
\setlength{\tabcolsep}{3pt}
\resizebox{\linewidth}{!}{%
\begin{tabular}{@{}p{0.1\linewidth}p{0.1\linewidth}p{0.3\linewidth}p{0.25\linewidth}p{0.12\linewidth}@{}}
\toprule
Benchmark & Scenario & Semantic request \(\varphi_f\) & Retain behavior & Backbone \\
\midrule
SMACv2 \citep{ellis2023smacv2} & Zerg 5v5 & Role-invariant focus-fire: two or more allied units attack the same enemy at the same decision step, independent of ally identities. & Episode return, battle win rate, and killed-enemy count. & MAPPO~\citep{yu2022surprising} \\
\midrule
Overcooked-AI/OGC \citep{carroll2019utility,ruhdorfer2024overcooked} & Asymmetric Advantages & Fixed cook--deliver split: one player contributes cooking/potting events while the partner contributes dish, soup, or delivery events. & Sparse/shaped soup return and total cooking plus delivery event count. & IPPO~\citep{de2020independent} \\
\midrule
RWARE \citep{papoudakis2020benchmarking} & Tiny 2-agent warehouse & Fixed carrier--yielder aisle convention: the target carrier handles shelf deliveries while the paired robot yields during pickup or carrying. & Shelf-delivery return and non-target delivery count. & QMIX~\citep{rashid2020monotonic} \\
\midrule
LBF \citep{papoudakis2020benchmarking} & 8x8-2p & Fixed-pair collection: the target pair jointly collects sparse-reward food. & Food-collection return and non-target pair collections. & VDN~\citep{sunehag2017value} \\
\midrule
VMAS \citep{bettini2022vmas} & GiveWay & Yield-then-cross: at the narrow crossing, one agent stays low-motion near the center while another crosses with high motion. & Task return, completion rate, and collision avoidance. & MADDPG~\citep{lowe2017multi} \\
\midrule
VMAS \citep{bettini2022vmas} & Passage & Single-file gate priority: near the passage, one agent queues or yields while another takes priority through the bottleneck. & Passage return, completion rate, and collision avoidance. & MAPPO~\citep{yu2022surprising} \\
\midrule
VMAS \citep{bettini2022vmas} & Flocking & Leader-follower dominance: a cohesive flock moves while one agent has a disproportionately large action norm and drives the group. & Flocking return, completion rate, and collision avoidance. & IPPO~\citep{de2020independent} \\
\bottomrule
\end{tabular}
}
\label{tab:experiment-suite}
\end{table*}
\renewcommand{\algorithmicrequire}{ \textbf{Input:}}     
\renewcommand{\algorithmicensure}{ \textbf{Output:}}    

\section{Algorithm Details}
\subsection{\CARVE{} Pseudocode}
\label{app:carve-pseudocode}
As shown in Algorithm \ref{alg:carve}, \CARVE{} is a post-training unlearning framework. It starts from a restored cooperative policy, uses the automaton monitor to locate decisions that causally support the forbidden convention, restricts the update to retain-aware strategy-sensitive directions, distills replacement coordination, and finally audits the edited actor under stronger elicitation contexts. The automaton and red-team contexts are used during editing and verification; the delivered policy is the edited actor itself and does not require an inference-time shield.

\subsection{Complexity Analysis}
Let \(M\) be the number of collected rollouts, \(H\) the horizon, \(N\) the number of agents, \(P\) the number of actor parameters, \(C\) the number of sampled coalitions per target-positive decision, \(K\) the retained subspace rank, \(f\) the editable actor-parameter fraction, \(U\) the number of projected optimization updates, \(R\) the number of audit contexts, and \(E\) the number of audit episodes per context. Monitor annotation is linear in trajectory length, \(O(MH)\), excluding simulator cost. Attribution requires \(O(MHNC)\) semantic-critic evaluations in the sampled-coalition implementation, rather than enumerating all \(2^N\) coalitions. Building the retain-aware subspace costs \(O(MHNP)\) to accumulate score-gradient statistics plus \(O(I_{\mathrm{eig}}KP)\) for \(I_{\mathrm{eig}}\) iterative generalized-eigen updates when diagonal or low-rank sensitivities are used; a dense \(P\times P\) eigensolver would be \(O(P^3)\) and is not the intended implementation. Selecting the edit mask costs \(O(P)\) to score coordinates plus a top-\(\lceil fP\rceil\) selection. Projected optimization costs \(O(U B_{\mathrm{mb}} P)\) for minibatch actor forward/backward passes, but only \(fP\) actor coordinates receive nonzero updates. Materializing the masked subspace costs \(O(KfP)\). The final audit costs \(O(REH)\) simulator steps. In practice, the dominant costs are simulator rollouts and actor-gradient passes; memory is \(O(P+KfP)\) for the actor and masked subspace, plus the rollout buffer.

\begin{algorithm}[t!]
\caption{\CARVE{} semantic coordination-strategy unlearning.}
\label{alg:carve}
\small
\begin{algorithmic}[1]
\REQUIRE Restored policy \(\pi_0\) with actor parameters \(\theta_0\);
simulator \(\mathcal G\); forbidden formula \(\varphi_f\);
retain data \(\mathcal D_r\);
monitor automaton \(\mathcal A_{\varphi_f}\);
edit rank \(K\); actor edit fraction \(f\);
update budget \(U\);
audit contexts \(\mathcal C_{\mathrm{audit}}\);
minimum semantic reduction \(\delta>0\);
utility threshold \(\rho\);
required consecutive passes \(M\);
audit interval \(A\).
\ENSURE Accepted edited policy \(\pi^-\), or \textsc{NotAccepted}.

\STATE Evaluate baseline scores
\(S_0=\mathrm{SSR}(\pi_0)\) and
\(W_0=\mathrm{WER}(\pi_0)\)
on \(\mathcal C_{\mathrm{audit}}\).

\STATE Roll out \(\pi_0\) in training and retain contexts;
update product states \(\bar s_t=(s_t,q_t)\)
with \(\mathcal A_{\varphi_f}\).
\STATE Mark target-positive and retain decision points
from the monitored trajectories.
\STATE Fit \(Q^\varphi(\bar s_t,\mathbf a_t)\)
for target-convention risk and
\(Q^J(\bar s_t,\mathbf a_t)\) for task value.

\FOR{each target-positive decision point}
    \STATE Sample coalitions and estimate
    \(\chi_{i,t}\), \(\chi^{(2)}_{ij,t}\)
    by Eqs.~\eqref{eq:individual-attribution}
    and~\eqref{eq:pairwise-attribution}.
    \STATE Convert attribution scores into weights \(w_{i,t}\)
    by Eq.~\eqref{eq:attribution-weight}.
\ENDFOR

\STATE Compute \(g_f(\tau)\)
by Eq.~\eqref{eq:forbidden-direction};
estimate \(G_f\) and \(F_r\).
\STATE Solve Eq.~\eqref{eq:retain-aware-eigenproblem};
retain the top-$K$ candidate directions $B=[u_1,\ldots,u_K]$.
\STATE Select the top $\lceil fP\rceil$ coordinates by retain-aware semantic sensitivity; form $m_f$ and $B_f=\operatorname{diag}(m_f)B$ as a heuristic sparsification.
\STATE Fix $B_f$ and restrict subsequent optimization to $\theta_\alpha=\theta_0+B_f\alpha$.
\STATE Build the centralized replacement teacher \(\widetilde\mu_t\) on high-risk product states using Eq.~\eqref{eq:cf-teacher}.
\STATE Define the decentralized distillation loss
\(\mathcal L_{\mathrm{cf}}\) by Eq.~\eqref{eq:cf-loss}.

\STATE Initialize \(\alpha\leftarrow 0\).
\FOR{\(u=1,\ldots,U\)}
    \STATE Update \(\alpha\) with Eq.~\eqref{eq:projection-objective},
    using \(\mathcal L_{\mathrm{retain}}\)
    from Eq.~\eqref{eq:retain-loss}.

    \IF{\(u\bmod A=0\) \OR \(u=U\)}
        \STATE Audit \(\pi_{\theta_\alpha}\)
        on \(\mathcal C_{\mathrm{audit}}\);
        obtain \(S_u=\mathrm{SSR}(\pi_{\theta_\alpha})\),
        \(W_u=\mathrm{WER}(\pi_{\theta_\alpha})\),
        and \(V_u=\mathrm{Util}(\pi_{\theta_\alpha})\).

        \IF{\(S_0-S_u\ge\delta\)
        \AND \(W_0-W_u\ge\delta\)
        \AND \(V_u\ge\rho\)}
            \STATE Accept \(\alpha^\star\leftarrow\alpha\);
            merge \(\Delta\theta=B_f\alpha^\star\)
            into the actors.
            \STATE Report SSR/WER,
            retention metrics, and ParamFrac.
            \RETURN \(\pi^-=\pi_{\theta_{\alpha^\star}}\).
        \ENDIF
    \ENDIF
\ENDFOR
\RETURN \textsc{NotAccepted}.
\end{algorithmic}
\end{algorithm}

\section{Evaluation and Hyperparameters Details}
\label{sec:a_D_Details}
Table~\ref{tab:experiment-suite} summarizes the task scenarios, role-invariant semantic unlearning requests, retained behaviors, and backbone algorithms across five cooperative MARL environments. 
Unless otherwise stated, SMACv2 and VMAS backbone training uses five random seeds, \(3.072\)M environment frames per run, 32 evaluation episodes every \(122{,}880\) frames, and checkpointing every \(307{,}200\) frames. SMACv2 uses four environment workers and 2,000 collected frames per batch, while VMAS uses 32 workers and 6,144 collected frames per batch. 
For post-checkpoint \CARVE{} runs, the default setting is 800 update steps, actor edit fraction \(f=0.12\), retain-action weight \(\lambda_r=0.5\), learning rate \(2\times10^{-4}\), \(\beta=1\), \(K=8\), \(\lambda_0=10^{-3}\) and \(\lambda=8\). For deterministic actors such as MADDPG, candidate actions are generated by adding independent Gaussian noise $\varepsilon\sim\mathcal N(0,0.15^2I)$ in the $\operatorname{atanh}$ latent space of normalized original actions, followed by a $\tanh$ transformation and rescaling to the native action range.

\section{Parameter Sensitivity}
\label{sec:para}
We sweep the hyperparameters that control the erasure--retention frontier on the three VMAS post-checkpoint tasks: GiveWay, Passage, and Flocking. 
The validation grid is update steps in \{200, 400, 800, 1200\}, edit fraction in \{0.03, 0.06, 0.12, 0.24, 0.48\}, retain weight in \{0.1, 0.5, 1.0, 2.0, 4.0\}, and learning rate in \{$10^{-4}$, $2\times10^{-4}$, $5\times10^{-4}$\}. Table~\ref{tab:sensitivity-unlearning} reports the average semantic reduction, retained utility, task reward, and parameter locality together. Increasing the edit fraction from \(0.06\) to \(0.12\) improves \(\Delta\)WER by \(16.7\%\), whereas expanding it to \(0.48\) yields only a further \(1.5\%\) gain with four times as many changed parameters. Increasing the retain weight from \(0.5\) to \(4.0\) reduces \(\Delta\)WER by \(44.6\%\), with little additional utility retention. Among the tested learning rates, \(2\times10^{-4}\) achieves the strongest semantic reductions. More updates do not consistently improve forgetting: \(800\) steps increase mean reward by \(50.1\%\) over \(200\) steps with comparable semantic reductions, while \(1200\) steps weaken forgetting. These results support the default configuration as a practical balance between convention removal, task performance, and edit locality.

\begin{table}[t!]
\centering
\caption{Parameter sensitivity on VMAS.}
\resizebox{\linewidth}{!}{\begin{tabular}{llrrrrr}
\toprule
\multirow{2}{*}{Sweep} & \multirow{2}{*}{Value} & \multicolumn{2}{c}{Unlearning} & \multicolumn{3}{c}{Retention/locality} \\
\cmidrule(lr){3-4}\cmidrule(lr){5-7}
 & & $\Delta$SSR$\uparrow$ & $\Delta$WER$\uparrow$ & Util.$\uparrow$ & Reward$\uparrow$ & ParamFrac$\downarrow$ \\
\midrule
Edit fraction & 0.03 & 0.122 & 0.253 & 0.966 & 49.1 & 0.030 \\
Edit fraction & 0.06 & 0.164 & 0.336 & 0.999 & 52.8 & 0.060 \\
Edit fraction & 0.12 & 0.173 & 0.392 & 0.999 & 54.2 & 0.120 \\
Edit fraction & 0.24 & 0.176 & 0.398 & 0.999 & 54.3 & 0.240 \\
Edit fraction & 0.48 & 0.181 & 0.398 & 0.999 & 54.8 & 0.480 \\
\midrule
Retain weight & 0.1 & 0.187 & 0.395 & 0.977 & 52.1 & 0.120 \\
Retain weight & 0.5 & 0.173 & 0.392 & 0.999 & 54.2 & 0.120 \\
Retain weight & 1.0 & 0.170 & 0.387 & 0.999 & 53.0 & 0.120 \\
Retain weight & 2.0 & 0.164 & 0.278 & 0.999 & 52.8 & 0.120 \\
Retain weight & 4.0 & 0.118 & 0.217 & 1.000 & 50.9 & 0.120 \\
\midrule
Learning rate & 1e-4 & 0.134 & 0.338 & 0.965 & 52.8 & 0.120 \\
Learning rate & 2e-4 & 0.173 & 0.392 & 0.999 & 54.2 & 0.120 \\
Learning rate & 5e-4 & 0.144 & 0.377 & 0.998 & 53.3 & 0.120 \\
\midrule
Update steps & 200 & 0.177 & 0.399 & 1.000 & 36.1 & 0.120 \\
Update steps & 400 & 0.137 & 0.318 & 0.998 & 53.5 & 0.120 \\
Update steps & 800 & 0.173 & 0.392 & 0.999 & 54.2 & 0.120 \\
Update steps & 1200 & 0.130 & 0.335 & 0.997 & 56.7 & 0.120 \\
\bottomrule
\end{tabular}
}
\label{tab:sensitivity-unlearning}
\end{table}

\section{Teacher--Actor Gap Experiment}
Marginal distillation may discard joint-action correlations encoded by the centralized teacher. Table~\ref{tab:vmas-teacher-actor-gap} therefore compares the joint teacher $T$ with the independently sampled edited actor $A_s$ on paired nominal episodes, using one trained checkpoint per VMAS scenario. Both are nominal-context score diagnostics. The observed gap is scenario-dependent. GiveWay shows nearly unchanged semantic scores and return. In Passage, the actor's SSR and WER point estimates decrease by $0.023$ and $0.022$, respectively, alongside a higher return point estimate. In Flocking, they increase by $0.019$ and $0.039$, respectively, alongside a lower return point estimate. Thus, the actor is not uniformly worse than the teacher. The teacher's local, critic-based KL-regularized objective does not guarantee dominance on episode-level metrics. All paired bootstrap 95\% confidence intervals for $J(A_s)-J(T)$ contain zero, leaving return differences statistically unresolved. This does not establish equivalence, particularly given the wide Passage interval. The comparison quantifies the combined effects of correlation loss, local observability, and constrained actor fitting.

\begin{table}[t!]
\centering
\caption{Joint-teacher versus actor performance on three VMAS scenarios. WER is measured in the nominal context. Confidence intervals bootstrap paired evaluation episodes.}
\setlength{\tabcolsep}{4pt}
\resizebox{\linewidth}{!}{
\begin{tabular}{@{}lrrrrrrc@{}}
\toprule
Scenario & \multicolumn{2}{c}{SSR$\downarrow$} & \multicolumn{2}{c}{WER$\downarrow$} & \multicolumn{2}{c}{Return$\uparrow$} & Paired 95\% CI \\
\cmidrule(lr){2-3}\cmidrule(lr){4-5}\cmidrule(lr){6-7}
 & $T$ & $A_s$ & $T$ & $A_s$ & $T$ & $A_s$ & of $J(A_s)-J(T)$ \\
\midrule
GiveWay & 0.312 & 0.313 & 0.315 & 0.313 & 3.678 & 3.679 & $[-0.0038,\,0.0056]$ \\
Passage & 0.231 & 0.208 & 0.362 & 0.340 & 121.1 & 158.2 & $[-8.098,\,230.861]$ \\
Flocking & 0.138 & 0.157 & 0.183 & 0.222 & 0.604 & 0.557 & $[-0.286,\,0.035]$ \\
\bottomrule
\end{tabular}}
\label{tab:vmas-teacher-actor-gap}
\end{table}

\section{Theory}
We analyze attribution, surrogate subspace selection, the replacement teacher, conditional return retention, and decentralized information limits.

\paragraph{Attribution consistency.}
\begin{proposition}[Shapley efficiency and estimation error]
\label{prop:theory-attribution}
For a fixed intervention baseline, let \(\chi_{i,t}\) and \(\widehat\chi_{i,t}\) be the exact Shapley values of \(v_t\) and its estimate \(\widehat v_t\). If \(\max_S|\widehat v_t(S)-v_t(S)|\le\varepsilon_v\), then
\begin{equation}
\sum_i\chi_{i,t}=v_t(\mathcal N)-v_t(\emptyset),
\qquad
|\widehat\chi_{i,t}-\chi_{i,t}|\le2\varepsilon_v.
\end{equation}
The pair term in Eq.~\eqref{eq:pairwise-attribution} vanishes for additive \(v_t\). These statements concern signed attributions before clipping.
\end{proposition}

\begin{proof}
Shapley values average permutation marginals, whose sum telescopes from \(v_t(\emptyset)\) to \(v_t(\mathcal N)\). Each estimated marginal has error at most \(2\varepsilon_v\). Additive coalition functions have zero second differences.
\end{proof}

\paragraph{Retain-aware subspace.}
Let \(G_f,F_r\succeq0\) and \(M=F_r+\lambda_0I\), with \(\lambda_0>0\). The semantic sensitivity surrogate is
\begin{equation}
\Delta\theta^\top G_f\Delta\theta
=\mathbb E[(g_f(\tau)^\top\Delta\theta)^2].
\label{eq:semantic-sensitivity-surrogate}
\end{equation}

\begin{theorem}[Retain-normalized sensitivity]
\label{thm:theory-subspace}
For \(1\le K\le d\), the top \(K\) generalized eigenvectors of \(G_fu=\lambda Mu\), normalized by \(B^\top MB=I_K\), solve
\begin{equation}
\max_{B\in\mathbb R^{d\times K}}\operatorname{tr}(B^\top G_fB)
\quad\mathrm{s.t.}\quad B^\top MB=I_K.
\end{equation}
The optimum is the sum of the top \(K\) generalized eigenvalues.
\end{theorem}

\begin{proof}
Set \(Y=M^{1/2}B\) and \(A=M^{-1/2}G_fM^{-1/2}\). The problem becomes \(\max_{Y^\top Y=I_K}\operatorname{tr}(Y^\top AY)\); the Ky Fan maximum principle gives the result.
\end{proof}

Because Eq.~\eqref{eq:semantic-sensitivity-surrogate} is unchanged by \(\Delta\theta\mapsto-\Delta\theta\), this theorem does not guarantee semantic-risk reduction. For a fixed coordinate mask, it applies to an eigenproblem solved on the selected principal submatrices, with \(K\) no larger than the selected dimension; it does not imply optimality of the mask or of a subsequently masked unrestricted basis.

\paragraph{Replacement teacher.}
At a fixed context, write \(p_0(\mathbf a)=\pi_0(\mathbf a\mid\mathbf h_t)\) and \(w_t(\mathbf a)=\beta Q^J(\bar s_t,\mathbf a)-\lambda Q^\varphi(\bar s_t,\mathbf a)\).

\begin{proposition}[KL-regularized teacher]
\label{prop:theory-teacher}
For bounded \(w_t\), the teacher in Eq.~\eqref{eq:cf-teacher} uniquely minimizes
\begin{equation}
D_{\mathrm{KL}}(\mu\Vert p_0)-\mathbb E_\mu[w_t]
\end{equation}
over distributions \(\mu\ll p_0\).
\end{proposition}

\begin{proof}
With \(Z_t=\mathbb E_{p_0}[e^{w_t}]\in(0,\infty)\),
\begin{equation}
D_{\mathrm{KL}}(\mu\Vert p_0)-\mathbb E_\mu[w_t]
=D_{\mathrm{KL}}(\mu\Vert\widetilde\mu_t)-\log Z_t,
\end{equation}
which is minimized uniquely at \(\mu=\widetilde\mu_t\).
\end{proof}
This optimum concerns the centralized teacher; marginal distillation need not preserve its joint-action correlations.

\paragraph{KL-based retention.}
For \(\pi^-=\pi_{\theta_\alpha}\), define
\begin{equation}
\kappa_\alpha(\mathbf h_t)
=\sum_iD_{\mathrm{KL}}\!\left(\pi_{0,i}(\cdot\mid h_t^i)\Vert\pi^-_i(\cdot\mid h_t^i)\right),
\overline{\mathcal L}_{\mathrm{retain}}
=\sum_t\mathbb E_{q_{r,t}}[\kappa_\alpha].
\label{eq:population-retain-kl}
\end{equation}
Here \(q_{r,t}\) is the population retain-history marginal, with the time weights of Eq.~\eqref{eq:retain-loss}; \(\mathcal L_{\mathrm{retain}}\) is its empirical counterpart.

\begin{proposition}[Conditional return retention]
\label{prop:theory-return}
Fix an \(H\)-step context \(c\) with common initial, transition, observation, and reward kernels, \(|r_t|\le R_{\max}\), and \(0\le\gamma\le1\). Let \(d_{\pi_0,t}^c\) be the original-policy history law. Suppose, for all \(t\),
\begin{equation}
d_{\pi_0,t}^c\le C_cq_{r,t},
\qquad
\overline{\mathcal L}_{\mathrm{retain}}(\alpha)
\le\mathcal L_{\mathrm{retain}}(\alpha)+\varepsilon_{\mathrm{gen}},
\label{eq:retain-history-coverage}
\end{equation}
where the first inequality is between measures, \(C_c<\infty\), and \(\varepsilon_{\mathrm{gen}}\ge0\) is finite and valid for the fitted edit. With \(R_H=R_{\max}\sum_{t=0}^{H-1}\gamma^t\),
\begin{equation}
|J_c(\pi^-)-J_c(\pi_0)|
\le2R_H\sqrt{\frac{C_c(\mathcal L_{\mathrm{retain}}(\alpha)+\varepsilon_{\mathrm{gen}})}{2}}.
\label{eq:conditional-return-retention}
\end{equation}
\end{proposition}

\begin{proof}
Common environment kernels and factorized actors give
\begin{equation}
D_{\mathrm{KL}}(P_{\pi_0}^c\Vert P_{\pi^-}^c)
=\sum_t\mathbb E_{d_{\pi_0,t}^c}[\kappa_\alpha]
\le C_c\overline{\mathcal L}_{\mathrm{retain}}
\le C_c(\mathcal L_{\mathrm{retain}}+\varepsilon_{\mathrm{gen}}).
\label{eq:retain-trajectory-kl-chain}
\end{equation}
Bounded returns and Pinsker's inequality yield
\(
|J_c(\pi^-)-J_c(\pi_0)|
\le2R_H\sqrt{D_{\mathrm{KL}}(P_{\pi_0}^c\Vert P_{\pi^-}^c)/2},
\)
proving the claim.
\end{proof}
The KL direction matches Eq.~\eqref{eq:retain-loss}. Coverage and generalization are additional conditions, not certified by a small replay loss or action-matching MSE; the experiments do not verify them.

\paragraph{Decentralized information limit.}
\begin{theorem}[Indistinguishable decision situations]
\label{thm:theory-decentralized}
If two same-time situations \(x,x'\) give every agent identical local histories but require disjoint joint-action sets \(\mathcal A_{\mathrm{good}}(x)\) and \(\mathcal A_{\mathrm{good}}(x')\), no history-conditioned decentralized policy satisfies both requirements with probability one. At least one situation has violation probability at least \(1/2\).
\end{theorem}

\begin{proof}
Identical histories induce the same joint-action law \(\mu\). Disjointness implies
\(
\mu(\mathcal A_{\mathrm{good}}(x))+\mu(\mathcal A_{\mathrm{good}}(x'))\le1,
\)
so at least one required set has probability at most \(1/2\).
\end{proof}
This theorem identifies an information limit that parameter editing alone cannot resolve. Its error bound is conditional on the two indistinguishable decision situations and does not directly bound overall SSR/WER or expected return.

\section{Limitations}
The formulation assumes cooperative finite-horizon MARL, CTDE training, target strategies expressible through observable relational predicates, and limited simulator access for suffix regeneration and red-team auditing. If the original reward continues to make the forgotten strategy uniquely optimal, subsequent unconstrained training can relearn it; preventing that is continual semantic revocation rather than one-shot unlearning. The proposed guarantees are behavioral and semantic, not direct legal-data-deletion guarantees. Future work will investigate replacement policies that better preserve joint-action correlations during decentralized distillation and broader auditing over partners and deployment contexts.
\end{document}